\documentclass{new_tlp}

\usepackage{url,graphicx}

\newcommand{\ignore}[1]{}

\newcommand\bcmdtab{\noindent\bgroup\tabcolsep=0pt%
  \begin{tabular}{@{}p{10pc}@{}p{20pc}@{}}}
\newcommand\ecmdtab{\end{tabular}\egroup}

  \title[Planning as Tabled Logic Programming]
        {Planning as Tabled Logic Programming}

  \author[N.-F. Zhou]
         {NENG-FA ZHOU\\
         CUNY Brooklyn College and Graduate Center \\
         \and Roman Bart\'{a}k \\
         Charles University \\
         \and Agostino Dovier \\
         Univ. di Udine \\
         }

\jdate{February 2014}
\pubyear{2014}
\pagerange{\pageref{firstpage}--\pageref{lastpage}}
\doi{XXX}

\begin{document}

\label{firstpage}

\maketitle

  \begin{abstract}
This paper describes Picat's planner, its implementation, and planning models for several domains used in International Planning Competition (IPC) 2014. Picat's planner is implemented by use of tabling. During search, every state encountered is tabled, and tabled states are used to effectively perform resource-bounded search. In Picat, structured data can be used to avoid enumerating all possible permutations of objects, and term sharing is used to avoid duplication of common state data. This paper presents several modeling techniques through the example models, ranging from designing state representations to facilitate data sharing and symmetry breaking, encoding actions with operations for efficient precondition checking and state updating, to incorporating domain knowledge and heuristics. Broadly, this paper demonstrates the effectiveness of tabled logic programming for planning, and argues the importance of modeling despite recent significant progress in domain-independent PDDL planners.
\end{abstract}

\section{Introduction}
Planning and logic programming are two close areas of research. PLANNER \cite{Hewitt69}, which was designed as \emph{a language for proving theorems and manipulating models in a robot}, is perceived as the first logic programming language. Planning has been an important problem domain for Prolog \cite{Kowalski79,warplan}. Despite the amenability of Prolog to planning, Prolog is no longer a competitive tool for planning.

Tabling \cite{Michie68,Tamaki86,warren92} is a technique used in logic and functional programming systems, which caches the results of certain calculations in memory and reuses them in subsequent calculations through a quick table lookup. Like state marking used in search algorithms, tabling can prevent the same state from being expanded more than once during search. Tabling has been found useful in many search problems, including theorem proving \cite{Nielson04,Pientka03}, program analysis \cite{Dawson96} and  model checking \cite{RamakrishnanRRSSW97}. Recently, tabled logic programming has been successfully employed to solve specific planning problems \cite{BartakZ14,ZhouD13,Zhou14}, and has been shown to be significantly faster than the state-of-the-art ASP (Answer Set Programming) planners on some problems \cite{ZhouD13,Zhou14}.

This paper describes Picat's planner and its implementation. This paper also presents planning models in Picat for several domains used in International Planning Competition 2014 (IPC'14) \cite{IPC14}, and demonstrates broader applicability of tabled logic programming to planning. Picat is a logic-based multi-paradigm language that provides logic variables, pattern matching, nondeterminism through backtracking, loops, functions, constraints, and tabling as its core modeling and solving features. As a modeling language for planning, Picat differs from PDDL (Plan Domain Description Language) \cite{PDDL} and ASP \cite{Brewka:2011:ASP,gekakasc12a,Lifschitz02} in several aspects: (1) Picat allows use of structures to represent states; (2) Picat supports explicit commitment and nondeterministic actions, which enables users to have better control over action applications; (3) Picat provides facilities for describing domain knowledge and heuristics for pruning search space. 

As a solving system, Picat's planner implements several techniques for better performance. First, it tables every state encountered during search and avoids repeating the exploration of the same state. Second, it adopts the hash-consing technique \cite{ZhouH12} to share common state data and to speed up the equality testing of states. Third, it utilizes tabled states to effectively perform {\it resource-bounded} search. For optimal planning, Picat offers built-ins to perform iterative search, but unlike IDA* \cite{Korf85}, Picat also reuses results tabled in early iterations \cite{Zhou14}.

This paper shows that the above-mentioned features of Picat make Picat a more appropriate language than PDDL for modeling and solving planning problems. To that end, this paper presents examples in Picat for several domains used in IPC'14. These examples illustrate several modeling techniques on how to design state representations to facilitate data sharing and symmetry breaking, on how to translate PDDL operators into Picat actions, and on how to incorporate domain knowledge and heuristics to reduce search spaces. This paper also gives the experimental results of the presented models and several other models encoded in the same way. The experimental results demonstrate the effectiveness of tabling and the importance of modeling. 

\section{A Brief Overview of Picat}
Picat is a dynamically-typed language. The basic types are taken from Prolog, except for {\it arrays} and {\it maps}. An array takes the form \texttt{\{$t_1$,$\ldots$,$t_{n}$\}}. The index of the first array element is 1, and the index notation {\tt X[I]} can be used to access array elements. Picat also borrows the basic logical operators from Prolog, including \textit{conjunction} ($A,B$), \textit{negation} ({\tt not $A$}), \textit{disjunction} ({\tt $A$;$B$}), and \textit{if-then-else} ({\tt $C$->$A$;$B$}).

Picat allows function calls in arguments. For this reason, it requires structures to be preceded with a dollar symbol \$ in order for them to be treated as data, unless the structure is special, or it occurs in a head pattern.

For each type, Picat provides a set of built-in functions and predicates. Many built-in predicates are taken from Prolog, including \texttt{member/2}, \texttt{nth/3}, and \texttt{select/3}. The function \texttt{insert\_ordered($List$,$Term$)} inserts $Term$ into the ordered list $List$ such that the resulting list remains ordered.

In Picat, predicates and functions are defined with pattern-matching rules. Picat has two types of rules: the {\it non-backtrackable rule} $Head, Cond\ $\verb+=>+$\ Body$, and the backtrackable rule $Head, Cond\ $\verb+?=>+$\ Body$. In a predicate definition, the $Head$ takes the form $p(t_1,\ldots,t_n)$, where $p$ is a predicate name, and $n$ is the arity. The condition $Cond$, which is an optional goal, specifies a condition under which the rule is applicable. For a call $C$, if $C$ matches $Head$ and $Cond$ succeeds, then the rule is said to be \emph{applicable} to $C$. When applying a rule to call $C$, Picat rewrites $C$ into $Body$. If the used rule is non-backtrackable, then the rewriting is a commitment, and the program can never backtrack to $C$. However, if the used rule is backtrackable, then the program will backtrack to $C$ once $Body$ fails, meaning that $Body$ will be rewritten back to $C$, and the next applicable rule will be tried on $C$. In a function definition, the $Head$ takes the form $f(t_1,\ldots,t_n) = Term$ where $f$ is a function name and $Term$ is a result to be returned. All of the rules in a function definition must be non-backtrackable.

A pattern can contain \emph{as-patterns} of the form \texttt{$V$@$Pattern$}, where $V$ is a new variable in the head, and $Pattern$ is a non-variable term. The as-pattern \texttt{$V$@$Pattern$} is the same as \texttt{$Pattern$} in pattern matching, but after pattern matching succeeds, $V$ is made to reference the term that matched $Pattern$. 

Picat supports loops and list comprehensions. For example, the loop 
\begin{verbatim}
    foreach(E in L) Goal end
\end{verbatim}
is true if {\tt Goal} is true for each {\tt E} in {\tt L}. Picat adopts the following simple scoping rule: {\it variables that occur only in a loop, but do not occur before the loop in the outer scope, are local to each iteration of the loop}. Loops are compiled into tail-recursive predicates, and list comprehensions are compiled into tail-recursive predicates through \texttt{foreach} loops.

Picat supports tabling for dynamic programming solutions. Other features of Picat include assignments, list comprehensions, global maps for storing permanent data, higher-order functions, action rules for defining event-driven actors, and modules for modeling and solving constraint satisfaction problems with CP, SAT, and MIP.

\section{Tabling in Picat}
Both predicates and functions can be tabled.  In order to have all calls and answers of a predicate or function tabled, users just need to add the keyword \texttt{table} before the first rule. For a predicate definition, the keyword {\tt table} can be followed by a tuple of table modes, including {\tt +} (input), {\tt -} (output), {\tt min}, {\tt max}, and {\tt nt} (not tabled). For a predicate with a table mode declaration that contains {\tt min} or {\tt max}, Picat tables one optimal answer for each tuple of the input arguments. The last mode can be \texttt{nt}, which indicates that the corresponding argument will not be tabled. 

Linear tabling \cite{Zhou08tab} is used in Picat, and table modes are taken from \cite{GuoG08}, except for the \texttt{nt} mode, which was initially proposed by \cite{Zhou10tai}.  Ground structured terms are hash-consed \cite{ZhouH12} so that common ground terms are tabled only once. For example, for the three lists \texttt{[1,2,3]}, \texttt{[2,3]}, and \texttt{[3]}, the shared sub-lists \texttt{[2,3]} and \texttt{[3]} are reused from \texttt{[1,2,3]}.

Mode-directed tabling has been successfully used to solve specific planning problems such as Sokoban \cite{ZhouD13}, and the Petrobras planning problem \cite{BartakZ14}. A planning problem is modeled as a path-finding problem over an implicitly specified graph. The following gives the framework used in all these solutions. 
\begin{small}
\begin{verbatim}
    table (+,-,min)
    path(S,Path,Cost), final(S) => Path = [],Cost = 0.
    path(S,Path,Cost) =>
        action(S,S1,Action,ActionCost),
        path(S1,Path1,Cost1),
        Path = [Action|Path1],
        Cost = Cost1+ActionCost.
\end{verbatim}
\end{small}
The call {\tt path(S,Path,Cost)} binds {\tt Path} to an optimal path from {\tt S} to a final state. The predicate {\tt final(S)} succeeds if {\tt S} is a final state, and the predicate {\tt action} encodes the set of actions in the problem. 

\begin{figure}[tb]
\begin{center}
\includegraphics[width=1.4in]{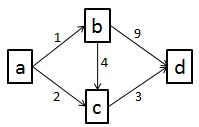}
\end{center}
\caption{\label{fig:shortest}A DAG.}
\end{figure}

Consider using the \texttt{path/3} predicate to find a shortest path from node \texttt{a} to node \texttt{d} in the DAG shown in Figure \ref{fig:shortest}. The final state is \texttt{d}. The call \texttt{path(a,Path,Cost)} initiates the search from node \texttt{a}. In order to resolve the call, Picat applies the transition \texttt{a$\rightarrow$b} to node \texttt{a}, and generates a new call to \texttt{path} for finding a shortest path from \texttt{b} to \texttt{d}. After the answer \texttt{b$\rightarrow$c$\rightarrow$d} is found for the call, Picat tables the answer \texttt{a$\rightarrow$b$\rightarrow$c$\rightarrow$d}, which has cost 8, for the inital call. After that, Picat backtracks, trying the alternative transition \texttt{a$\rightarrow$c} to node \texttt{a}. After the shortest path \texttt{c$\rightarrow$d} is found, Picat finds another path \texttt{a$\rightarrow$c$\rightarrow$d}, which has cost 5, for the initial call. Since this path is shorter than the tabled path, Picat replaces the tabled path with this new path. Since the graph has no cycles, Picat returns the new path as the final answer. In general, calls need to be re-evaluated until fixed points are reached if there are looping calls \cite{Zhou08tab}. 

When applied to the single-source shortest path problem, linear tabling is similar to Dijkstra's algorithm, except that linear tabling tables shortest paths from the encountered states to the goal state rather than shortest paths to the encountered states from the initial state. 

The above framework performs depth-unbounded search. For many planning problems, branch \& bound and IDA* \cite{Korf85} are useful for finding optimal solutions. The {\tt planner} module of Picat provides built-ins for planning with different types of search.

\section{The {\tt planner} Module of Picat}
The \texttt{planner} module is based on tabling but it abstracts away tabling from users. For a planning problem, users only need to define the predicates \texttt{final/1} and \texttt{action/4}, and call one of the search predicates in the module on an initial state in order to find a plan or an optimal plan.

\begin{itemize}
\item \texttt{final($S$)}: This predicate succeeds if $S$ is a final state.
\item \texttt{action($S$,$NextS$,$Action$,$ACost$)}: This predicate encodes the state transition diagram of a planning problem. The state $S$ can be transformed to $NextS$ by performing $Action$. The cost of $Action$ is $ACost$, which must be non-negative. If the plan's length is the only interest, then $ACost$ = 1.
\end{itemize}
These two predicates are called by the planner.  The {\tt action} predicate specifies the precondition, effect, and cost of each of the actions. This predicate is normally defined with nondeterministic pattern-matching rules. As in Prolog, the planner tries actions in the order they are specified. When a non-backtrackable rule is applied to a call, the remaining rules will be discarded for the call.

The following predicates constitute the core of the \texttt{planner} module.
\begin{itemize}
\item \texttt{best\_plan\_unbounded($S$,$Limit$,$Plan$,$PlanCost$)}: This predicate finds an optimal plan by performing {\it depth-unbounded} search. This predicate is implemented based on the path-finding framework shown above. The argument {\tt $Limit$} is not utilized to limit the depth of search. It is compared with $PlanCost$ after an optimal plan has been found. During depth-unbounded search, once a state has failed, it will not be explored again.

\item \texttt{plan($S$,$Limit$,$Plan$,$PlanCost$)}: This predicate searches for a plan by performing {\it resource-bounded} search, in which a state is expanded only if it is new and its resource limit is non-negative, or if the state has previously failed but the current occurrence has a higher resource limit than before. This predicate is defined as a tabled predicate that tables the $Limit$ argument but does not use it in variant checking \cite{Zhou14}. The implementation of this predicate is described in the next subsection.

\item \texttt{best\_plan($S$,$Limit$,$Plan$,$PlanCost$)}: This predicate finds an optimal plan by performing resource-bounded {\it iterative-deepening} search. It calls the {\tt plan/4} predicate to find a plan, using 0 as the initial cost limit and gradually relaxing the cost limit until a plan is found. Unlike IDA*, which starts a new round from scratch, Picat also reuses the states that were tabled in the previous rounds. 

\item \texttt{best\_plan\_bb($S$,$Limit$,$Plan$,$Cost$)} This predicate finds an optimal plan using branch \& bound. First, it calls \texttt{plan/4} to find a plan. Then, it tries to find a better plan by imposing a stricter limit. This step is repeated until no better plan can be found. It returns the last plan that was found.

\item {\tt current\_resource()}: This function returns the resource limit argument of the latest call to {\tt plan/4}. In order to retrieve the argument, the implementation has to traverse the call-stack until it reaches a call to {\tt plan/4}. This function can be used to check against a heuristic value. If the heuristic estimate of the cost to travel from the current state to a final state is greater than the resource limit, then the current state should fail.
\end{itemize}

\begin{figure}
\begin{center}
\begin{small}
\begin{verbatim}
    plan(S,Limit,Plan,PlanCost) =>
        IPlan = {Limit,[],0},
        catch(plan_bounded_aux(S,IPlan), (Plan,PlanCost), true).

    table (+,nt)
    plan_bounded_aux(S,{Limit,IPlan,IPlanCost}),
        final(S)
    =>
        throw((IPlan.reverse(), IPlanCost)).
    plan_bounded_aux(S,{Limit,IPlan,IPlanCost}) =>
        action(S,NextS,Action,ACost),
        Limit1 = Limit-ACost,
        Limit1 >= 0,
        Inherited1 = {Limit1,
                      [Action|IPlan],
                      IPlanCost+ACost},
        plan_bounded_aux(NextS,Inherited1).
\end{verbatim}
\end{small}
\end{center}
\caption{\label{fig:planbounded}The implementation of \texttt{plan/4}.}
\end{figure}

\section{The Implementation of Resource-Bounded Search}
Figure \ref{fig:planbounded} sketches Picat's implementation of the predicate \texttt{plan/4}. The following array is passed from a state to the next state as an \texttt{nt} argument: 
\begin{verbatim}
    {Limit,IPlan,IPlanCost}
\end{verbatim}
In order to perform resource-bounded search, Picat treats the second argument of the predicate \texttt{plan\_bounded\_aux} differently from other \texttt{nt} arguments. Picat stores the \texttt{Limit} argument of each failed call to \texttt{plan\_bounded\_aux}, and uses this information to decide whether the same state should fail when it recurs.

The first rule of \texttt{plan\_bounded\_aux} throws the inherited plan and its cost as an exception if \texttt{final(S)} succeeds. The exception will be caught by the \texttt{catch} call in the rule body of \texttt{plan/4}.

The second rule of \texttt{plan\_bounded\_aux} calls \texttt{action/4} to select an action, which produces a new state, \texttt{NextS}. Then, the rule computes the new resource limit, \texttt{Limit1}, by subtracting the cost of the selected action from \texttt{Limit}. If \texttt{Limit1 >= 0} succeeds, then the rule continues with the tabled search by recursively calling \texttt{plan\_bounded\_aux} on the new state \texttt{NextS}. Otherwise, if \texttt{Limit1 >= 0} fails, then Picat backtracks to select an alternative action.

\begin{figure}[tb]
\begin{center}
\includegraphics[width=.17\textwidth]{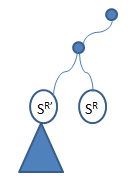}
\caption{\label{fig:resource_bounded} Resource-bounded search.}
\end{center}
\end{figure}

The idea of resource-bounded search is to utilize tabled states and their resource limits to effectively decide when a state should be expanded and when a state should fail. Let $S^R$ denote a state with an associated resource limit, $R$, as depicted in Figure \ref{fig:resource_bounded}. If $R$ is negative, then $S^R$ immediately fails. If $R$ is non-negative and $S$ has never been encountered before, then $S$ is expanded by using a selected action. Otherwise, if the same state $S$ has failed before and $R'$ was the resource limit when it failed, then $S^R$ is only expanded if $R>R'$, i.e., if the current resource limit is larger than the resource limit was at the time of failure.

\section{\label{sec:apps}Modeling Examples}
This section describes the Picat models for four of the problem domains used in the sequential optimal track of IPC'14: {\it Transport}, {\it Tetris}, {\it Floortile}, and {\it Parking}.  Each of the following models will show the state representation, the encoding of the actions, the search predicate that is used ({\tt best\_plan} or {\tt best\_plan\_unbounded}), and the domain knowledge and heuristics that are employed. In these models, states are typically represented by lists, and preconditions and state updates are handled by standard list operations. Sometimes, arrays are used when no list suffixes can be shared. The Picat encodings of actions are mostly straightforward translations from the PDDL encodings. These models do not use sophisticated domain knowledge or heuristics that would hurt the readability or compromise the optimality of answers. These four domains, as well as five other domains, are described in Appendix A.


\subsection{Example-1: Transport}
This problem is a variant of the popular logistics domain in planning. Given a weighted directed graph, a set of trucks each of which has a capacity for the number of packages it can carry, and a set of packages each of which has an initial location and a destination, the objective of the problem is to find an optimal plan to transport the packages from their initial locations to their destinations. This problem is more challenging than the {\it Nomystery} problem that was used in IPC'11, because of the existence of multiple trucks, and because an optimal plan normally requires trucks to cooperate. This problem degenerates into the shortest path problem if there is only one truck and only one package. The PDDL and Picat encodings of the problem are given in Appendix B.

\subsubsection*{Basic Encoding} 
A state is represented by an array of the form {\tt \{Trucks,Packages\}}, where {\tt Trucks} is an ordered list of trucks, and {\tt Packages} is an ordered list of waiting packages.  A package in {\tt Packages} is a pair of the form {\tt (Loc,Dest)} where {\tt Loc} is the source location and {\tt Dest} is the destination of the package. A truck in {\tt Trucks} is a list of the form {\tt [Loc,Dests,Cap]}, where {\tt Loc} is the current location of the truck, {\tt Dests} is an ordered list of destinations of the loaded packages on the truck, and {\tt Cap} is the capacity of the truck. At any time, the number of loaded packages must not exceed the capacity. 

Note that keeping {\tt Cap} as the last element of the list facilitates sharing, since the suffix {\tt [Cap]}, which is common to all the trucks that have the same capacity, is tabled only once. Also note that the names of the trucks and the names of packages are not included in the representation. Two packages in the waiting list that have the same source and the same destination are indistinguishable, and as are two packages loaded on the same truck that have the same destination. This representation breaks symmetries. Two configurations that only differ by a truck's name or a package's name are treated as the same state. 

A state is final if all of the packages have been transported.
\begin{small}
\begin{verbatim}
    final({Trucks,[]}) =>   
        foreach([_Loc,Dests|_] in Trucks)
            Dests == []
        end.
\end{verbatim}
\end{small}

The PDDL rules for the actions are straightforwardly translated into Picat as follows.
\begin{small}
\begin{verbatim}
    action({Trucks,Packages},NextState,Action,ACost) ?=>
        Action = $load(Loc), ACost = 1,
        select([Loc,Dests,Cap],Trucks,TrucksR),
        length(Dests) < Cap,
        select((Loc,Dest),Packages,PackagesR),   
        NewDests = insert_ordered(Dests,Dest),
        NewTrucks = insert_ordered(TrucksR,[Loc,NewDests,Cap]),
        NextState = {NewTrucks,PackagesR},
    action({Trucks,Packages},NextState,Action,ACost) ?=>
        Action = $unload(Loc), ACost = 1,    
        select([Loc,Dests,Cap],Trucks,TrucksR),
        select(Dest,Dests,DestsR),
        NewTrucks = insert_ordered(TrucksR,[Loc,DestsR,Cap]),
        NewPackages = insert_ordered(Packages,(Loc,Dest)),
        NextState = {NewTrucks,NewPackages}.
    action({Trucks,Packages},NextState,Action,ACost) =>
        Action = $move(Loc,NextLoc),
        select([Loc|Tail],Trucks,TrucksR),
        road(Loc,NextLoc,ACost),    
        NewTrucks = insert_ordered(TrucksR,[NextLoc|Tail]),
        NextState = {NewTrucks,Packages}.
\end{verbatim}
\end{small}
For the {\it load} action, the rule nondeterministically selects a truck that still has room for another package, and nondeterministically selects a package that has the same location as the truck. After loading the package to the truck, the rule inserts the package's destination into the list of loaded packages of the truck. Note that the rule is nondeterministic. Even if a truck passes by a location that has a waiting package, the truck may not pick it. If this rule is made deterministic, then the optimality of plans is no longer guaranteed, unless there is only one truck and the truck's capacity is infinite.  

\subsubsection*{Domain Knowledge and Heuristics} 
Domain knowledge can be used to reduce the nondeterminism and avoid unnecessary applications of actions. In the complete Picat encoding given in Appendix B, the predicate \texttt{action} begins with a rule that deterministically unloads a package if the package's destination is the same as the truck's location.

Resource-bounded search is used to find an optimal plan. After each new state is generated, the following condition is checked to ensure that the current path is viable.
\begin{small}
\begin{verbatim}
    current_resource() - ACost >=  estimated_cost(NewState).
\end{verbatim}
\end{small}
Let $P_1$, $\ldots$, $P_n$ be the remaining packages, and $C_i$ be the minimum cost of moving package $P_i$ to its destination by using any truck. The moving cost of a state can be safely estimated as {\tt max($\{C_1,\ldots,C_j\}$)}. The estimated total cost is the estimated moving cost plus the loading and unloading costs of all of the remaining packages. This heuristic is admissible. 

\subsection{Example-2: Tetris}
The problem is a simplified version of the well-known Tetris. There are three kinds of pieces: 1-cell {\it boxes}, 2-cell {\it rectangles}, and 3-cell {\it L-shaped} pieces.  Initially, all the pieces are distributed on a grid board. The pieces can freely move on the board to any cells as long as the cells are not occupied by other pieces. There is a region on the board that is designated as the target region. The goal of the game is to move all the pieces to the target region. In the IPC setting, rectangles can rotate but L-shaped pieces cannot.

\subsubsection*{Basic Encoding} 
Pieces can be represented as follows: A square is represented by the cell it occupies; a rectangle by a term of the form {\tt rect(C1,C2)} where {\tt C1} and {\tt C2} are cell locations; and an L-shaped piece by a term of the form {\tt ell(C1,C2,C3)}. In the IPC setting, the ordering of the cells that are occupied by a piece is important. So {\tt rect(C1,C2)} and {\tt rect(C2,C1)} represent two different rectangle pieces. A state is represented by an array of the form \texttt{\{Squares,Rects,Ls\}}, where each argument gives a sorted list of a single kind of pieces.

The PDDL rules for actions can be translated into Picat in a straightforward manner. For example, the following rule selects a rectangle piece to move.
\begin{small}
\begin{verbatim}
    action(State@{Squares,Rects,Ls},NewState,Action,ACost) ?=>
        Action = $move(Rect,NewRect), ACost = 1,
        select(Rect,Rects,RectsR),  
        Rect = $rect(C1,C2), 
        connected(C2,C3),
        not_occupied(C3,State),          %%  C3 is free
        NewRect = $rect(C2,C3), 
        NewState = {Squares,insert_ordered(Rects,NewRect),Ls}.
\end{verbatim}
\end{small}
The predicate {\tt not\_occupied(C3,State)} is true if {\tt C3} is not occupied by any of the pieces in {\tt State}.

\subsubsection*{Heuristics} 
Resource-bounded search is used to find an optimal plan for this problem. The goal of the problem is to move all the pieces to the target region. For each piece, assume that it's the only piece on the board, disregarding all other pieces. The estimated cost of moving the piece to the target region is its shortest distance to the nearest cell in the target region times the cost of each move. The estimated cost of transforming a state into a final state is the sum of the estimated costs of all the pieces. This heuristic function is admissible and easy to compute, but very conservative because a cell in the target region can serve as a target for multiple pieces.

\subsection{Example-3: Floortile}
A set of robots use two different colors (black and white) to paint patterns in floor tiles. The robots can move around the floor tiles in four directions (up, down, left and right). Robots paint with one color at a time, but can change their spray guns to the other color. However, robots can only paint the tile that is in front (up) and behind (down) them, and once a tile has been painted no robot can stand on it. 

\subsubsection*{Basic Encoding} 
A state is represented by a list of the form \texttt{[Robots,WTiles,BTiles]}, where {\tt Robots} is an ordered list of robots and {\tt WTitles} ({\tt BTiles}) is an ordered list of locations of the painted white (black) tiles. A robot in {\tt Robots} is a pair {\tt (Color,Loc)} where {\tt Color} is the color of the paint that the robot is holding and {\tt Loc} is the robot's location. 

A state is final if all the tiles that are required to be painted are all painted.
\begin{small}
\begin{verbatim}
    final({_,WTiles,BTiles}) => 
        painted_w_tiles_in_goal(WTiles),
        painted_b_tiles_in_goal(BTiles).
\end{verbatim}
\end{small}

The actions are encoded according to the state representation. For example, once a tile is painted, the tile's location is added into {\tt WTiles} or {\tt BTiles} depending on the color of the paint; once a robot {\tt (Color,Loc)} moves from {\tt Loc} to {\tt Loc1}, the pair is changed to {\tt (Color,Loc1)}. The current graph is determined by the initial graph, the set of painted tiles, and the set of robots. A robot can move from its current location {\tt Loc} to a new location {\tt NextLoc} if {\tt NextLoc} is connected to {\tt Loc} in the graph, {\tt NextLoc} is not painted, and {\tt NextLoc} is not occupied by another robot.

\subsubsection*{Macro Actions and Domain Knowledge} 
The color-changing action, if needed, can be forced to take place right before a painting action. This macro action helps reduce the search space.

For IPC'14, the used instances were generated in the fashion that {\it robots should only paint tiles in front of them}. This special condition can be exploited to reduce the nondeterminism of the painting rules: if a robot is at location {\tt Loc}, trying to paint the up location {\tt ULoc}, and the up location of {\tt ULoc} has already been painted, then {\tt ULoc} can be painted deterministically; similarly, if a robot is at location {\tt Loc}, trying to paint the down location {\tt DLoc}, and the down location of {\tt DLoc} has already been painted, then {\tt DLoc} can be painted deterministically.

Depth-unbounded search is used for this problem. During depth-unbounded search, dead ends are tabled and are not re-explored when they are encountered again. For this problem, a state becomes a dead end if there is an unpainted tile that is  not reachable by any robot. No extra code is needed to detect dead ends. Because depth-unbounded search is used, no heuristics are needed.

\subsection{Example-4: Parking}
This domain involves parking cars on a street with N curb locations, and where cars can be double-parked but not triple-parked. The goal is to find a plan to move from one configuration of parked cars to another configuration, by driving cars from one curb location to another. For each curb location, there are two parking spots: the curb side and the road side. For a car to move from a spot at a curb into a spot at a different curb, the destination spot must be clear; and if the spot is on the curb side, then the road-side spot must also be clear. In some ways, this problem is similar  to the {\it Blocks World} and the {\it Tower of Hanoi}.

\subsubsection*{Basic Encoding} 
A configuration is represented by an array of curbs $\{B_1,B_2,\ldots,B_n\}$, where each $B_i$ is a list of up to two cars. When a curb has two cars {\tt [$C_1$,$C_2$]}, it is assumed that $C_1$ is the road-side car and $C_2$ is the curb-side car. This representation allows the road-side car $C_1$ to be removed without touching the curb-side car $C_2$.

A state is represented by a pair of the form {\tt (CurrConfig,GoalConfig)}, where a {\tt CurrConfig} is the current configuration, and {\tt GoalConfig} is the goal configuration. As shown below, the inclusion of the goal configuration in the state representation facilitates the encoding of domain knowledge.

A state is final if the current configuration is the same as the goal configuration.
\begin{small}
\begin{verbatim}
    final((Config,Config)) => true.
\end{verbatim}
\end{small}

All possible moves are specified with one rule as follows:
\begin{small}
\begin{verbatim}
    action((Config,GConfig),NextS,Action,ACost) => 
        Action = $move(I,J), ACost = 1,
        nth(I,Config,ICars),ICars = [ClearCar|NewICars],
        nth(J,Config,JCars), J !== I, JCars != [_,_],
        N = length(Config),
        NewConfig = new_array(N),
        foreach (K in 1..N)
            (K == I -> NewConfig[K] = NewICars
            ;K == J -> NewConfig[K] = [ClearCar|JCars]
            ; NewConfig[K] = Config[K])
        end,
        NextS = (NextConfig,GConfig).
\end{verbatim}
\end{small}
The built-in predicate {\tt nth(I,Array,Arg)} is true if the {\tt I}-th argument of {\tt Array} is {\tt Arg}. When {\tt I} is a variable, this predicate nondeterministically searches for an argument that unifies with {\tt Arg}. The rule finds a non-empty curb {\tt I} and a different curb {\tt J} that has less than two cars, and moves the clear car, {\tt ClearCar}, from curb {\tt I} to curb {\tt J}. 

\subsubsection*{Domain Knowledge and Heuristics} 
Two knowledge rules can be incorporated into the basic encoding in order to speed up the search. First, when a car is moved into a spot that is its final spot in the goal configuration, such a move should be made deterministically. Note that if a car is moved to a road-side final spot then the curb-side spot must be occupied by a correct car. Second, a car that has been placed in its final spot should not be moved away. This kind of domain knowledge has been used in solving the Tower of Hanoi \cite{AlfordKN09} and the Blocks World \cite{BacchusK00}.

Resource-bounded search is used to find an optimal plan. The cost of transforming a state to a final state can be simply estimated as the number of incorrectly-positioned cars.  This estimation is admissible, since at least one action is needed to move each incorrectly-positioned car. The following improved heuristic function, which takes special curb configurations into account, is used by the model: For each curb, if the curb has two cars {\tt [A,B]} in the current state, but is required to have {\tt [B,A]} in the final state, then its cost is 4; if the curb has two cars {\tt [A,B]}, but is required to have {\tt [C,A]} or {\tt [B,C]} ({\tt A} $\neq$ {\tt C}, {\tt B} $\neq$ {\tt C}), then its cost is 3; otherwise, the cost is the number of incorrectly-positioned cars.

\section{\label{sec:results}Experimental Results}
In addition to the four domains presented in this paper, we have encoded in Picat several other domains used in the deterministic sequential track of IPC'14. The domains and their Picat encodings are given in Appendix A. The Picat encodings are simple, compact, and comparable in size with the PDDL encodings used in IPC'14. Most of the encodings can be further improved by incorporating sophisticated domain knowledge and heuristics. In order to evaluate the effectiveness of the use of tabling and the use of heuristics, we have built two separate sets of encodings, namely, {\sf Picat-nt} and {\sf Picat-nh}, which use the same state representation as the original Picat encodings but have some component removed. {\sf Picat-nt} performs Prolog-style non-tabled iterative-deepening search. Since {\sf Picat-nt} does not table any state, it may explore the same state multiple times during search. The {\sf Picat-nh} encodings do not use any heuristics. We have compared these Picat encodings with the IPC'14 PDDL encodings solved with {\sf Symba} \cite{Torralba14}, a domain-independent bidirectional A* planner which won the optimal sequential track of IPC'14. This comparison offers a glimpse of how well Picat compares with the best domain-independent planner. A comparison of Picat's planner and several domain-dependent planners also shows the promise of tabled planning \cite{RomanPPDP15}.

\begin{table}[t]
\caption{\label{tab:res}The number of instances solved optimally.}
\centering
\begin{small}
\begin{tabular}{|c|r|r|r|r|r|}  \cline{1-6} 
{\sf Domain}          &  {\sf \# insts} & {\sf Picat}    & {\sf Picat-nt} & {\sf Picat-nh} &  {\sf Symba} \\ \cline{1-6}
{\it Barman}          &    14     &   \textbf{14}      &     0    &   \textbf{14}     &   6      \\ \cline{1-6}
{\it Cave}            &    20     &   \textbf{20}      &     0    &    \textbf{20}    &   3      \\ \cline{1-6}
{\it Childsnack}      &    20     &   \textbf{20}      &    \textbf{20}    &   \textbf{20}     &   3     \\ \cline{1-6}
{\it Citycar}         &    20     &   \textbf{20}      &    17    &   18     &  17    \\ \cline{1-6}
{\it Floortile}       &    20     &   \textbf{20}      &     0    &   \textbf{20}     &  \textbf{20}    \\ \cline{1-6}
{\it GED}             &    20     &   \textbf{20}      &    19    &   13     &  19    \\ \cline{1-6}
{\it Parking}         &    20     &   \textbf{11}      &     4    &   0      &   1    \\ \cline{1-6}
{\it Tetris}          &    17     &   \textbf{13}      &    \textbf{13}    &   9       &  10   \\ \cline{1-6}
{\it Transport}       &    20     &    \textbf{9}      &     0    &  4       &   8    \\ \cline{1-6}
\end{tabular}
\end{small}
\end{table}

Table \ref{tab:res} shows the number of instances ({\sf \#insts}) in the domains used in IPC'14 and the number of (optimally) solved instances by each planner. The results were obtained on a Cygwin notebook computer with 2.4GHz Intel i5 and 4GB RAM. Both Picat and Symba were compiled using g++ version 4.8.3. For Symba, a setting suggested by one of Symba's developers was used. A time limit of 30 minutes was used for each instance as in IPC. For every instance that was solved by both Symba and Picat, the plan quality is the same.

A comparison of {\sf Picat} and {\sf Picat-nt} shows the effectiveness of the use of tabling. For every domain, except for {\it Childsnack} and {\it Tetris}, {\sf Picat} solved more instances than {\sf Picat-nt}. For {\it Barman}, {\it Cave}, {\it Floortile} and {\it Transport},  {\sf Picat-nt} could not solve any of the instances. The Picat encodings for five of the domains ({\it Citycar}, {\it GED}, {\it Parking}, {\it Tetris}, and {\it Transport}) use heuristics. The use of heuristics is helpful for these domains, especially for {\it Parking}, for which {\sf Picat-nh} did not solve any of the instances.

{\sf Picat} solved more instances than {\sf Symba} for every domain except for {\it Floortile}, for which both systems solved all of the instances. The running times of the instances are not given, but the total runs for {\sf Picat} were finished within 24 hours, while the total runs for {\sf Symba} took more than 72 hours.

\section{\label{sec:con}Conclusion and Future Work}
This paper has presented Picat's planner, its implementation, and example models for several domains from IPC'14. The example models illustrate several modeling techniques in Picat. One key task of modeling is finding an efficient state representation. While classical planning frameworks such as PDDL are based on a factored representation of states, Picat uses a structured representation. A structured state representation can leave out unnecessary information that is not needed for planning and can break symmetries by avoiding enumerating all possible permutations of objects.  Another key task of modeling is utilizing domain knowledge to reduce search spaces. In the past, a lot of work has been done on the use of domain knowledge in planning \cite{BacchusK00,Haslum03domainknowledge,kautz-aips98}, but recently, this part of modeling has been put aside, because of the advancement of domain-independent PDDL planners. This paper has shown that, even with simple domain knowledge, the declarative encodings of Picat significantly outperform Symba, a state-of-the-art domain-independent planner. 

This paper has demonstrated for the first time that tabled logic programming is competitive with the cutting-edge PDDL planners. The key to the success is tabling. Tabling avoids repeating the exploration of the same state and facilitates performing resource-bounded search. The Picat planner does not do prior grounding that is typical for most current state-space search planners, and hence Picat has no problem with exploded memory consumption due to grounding. Nevertheless, memory consumption can be demanding during search since every encountered state is tabled. This is why careful modeling that removes symmetries and uses domain control knowledge to prune useless state transitions is important.

A good state representation should also exploit the underlying term-sharing technique that is used in the tabling system. In the examples that are presented in this paper, ordered lists are used to represent collections of objects. It takes linear time to perform the basic operations. One direction for future work is to design data structures for state representations which are compact, efficient, and good for sharing. 

A plethora of action languages have been designed for modeling and solving planning problems (e.g., \textsf{\emph{A}} and its successors \cite{GelfondL98}, Golog \cite{LevesqueRLLS97}, and \textsf{\emph{K}} \cite{EiterFLPP04}). The focus of these languages has been on the modeling power rather than efficiency and scalability. These languages have been implemented by translation into SAT, ASP, CP, or PDDL \cite{bai-fri-mci-hecfest11,DFP11}, but no implementation has been shown to be competitive with the cutting-edge PDDL planners. Picat can be used as an implementation language for these action languages. Like in PDDL, a state is represented as a set of flat facts in all of these action languages. Another direction for future work is to devise an efficient translation from these action languages into Picat that automatically exploits structural representation, symmetries, domain control knowledge, and heuristics.

\clearpage

\clearpage
\appendix
\section{Benchmarks used in the paper}
In this section we summarize, for reader's convenience, 
the descriptions of all the domains used as benchmarks.
Descriptions are drawn from
\url{https://helios.hud.ac.uk/scommv/IPC-14/domains_sequential.html};
Picat's complete encodings for these benchmarks are available at \url{http://picat-lang.org/ipc14/}.

\subsection{Barman}

There is a robot \emph{barman} that manipulates drink dispensers, glasses, and a shaker. 
The goal is to find a plan of robot's actions that serves a desired set of drinks. 
Robot hands can grasp at most one object at a time.
Glasses need to be empty and clean to be filled.
The benchmark was proposed by  Sergio Jim\'enez Celorrio.

\begin{figure}[htb]
\begin{center}
\begin{tabular}{cc}
{\tt \small 
\begin{tabular}[b]{llll}
\%\% INITIAL state\\
ontable(shaker1),  & ontable(shot1),  \\
  ... , &  ontable(shot8),\\
clean(shaker1),  & clean(shot1), \\
 ... , &  clean(shot8),\\
empty(shaker1), & empty(shot1), \\
  ..., &  empty(shot8),\\
dispenses(dispenser1,ingredient1),  & 
dispenses(dispenser2,ingredient2),  \\ 
dispenses(dispenser3,ingredient3),  &
dispenses(dispenser4,ingredient4),\\
handempty(left),  & handempty(right), \\
\%\% Cocktail rules\\
cocktail\_part1(cocktail1,ingredient1),  & cocktail\_part2(cocktail1,ingredient3), \\
...\\
cocktail\_part1(cocktail6,ingredient2), & cocktail\_part2(cocktail6,ingredient1), \\
\%\% GOAL\\
contains(shot1,cocktail1),           &    contains(shot2,cocktail1), \\
contains(shot3,cocktail2),           &    contains(shot4,cocktail6).
 \end{tabular}
 }
\\
\includegraphics[width=4cm]{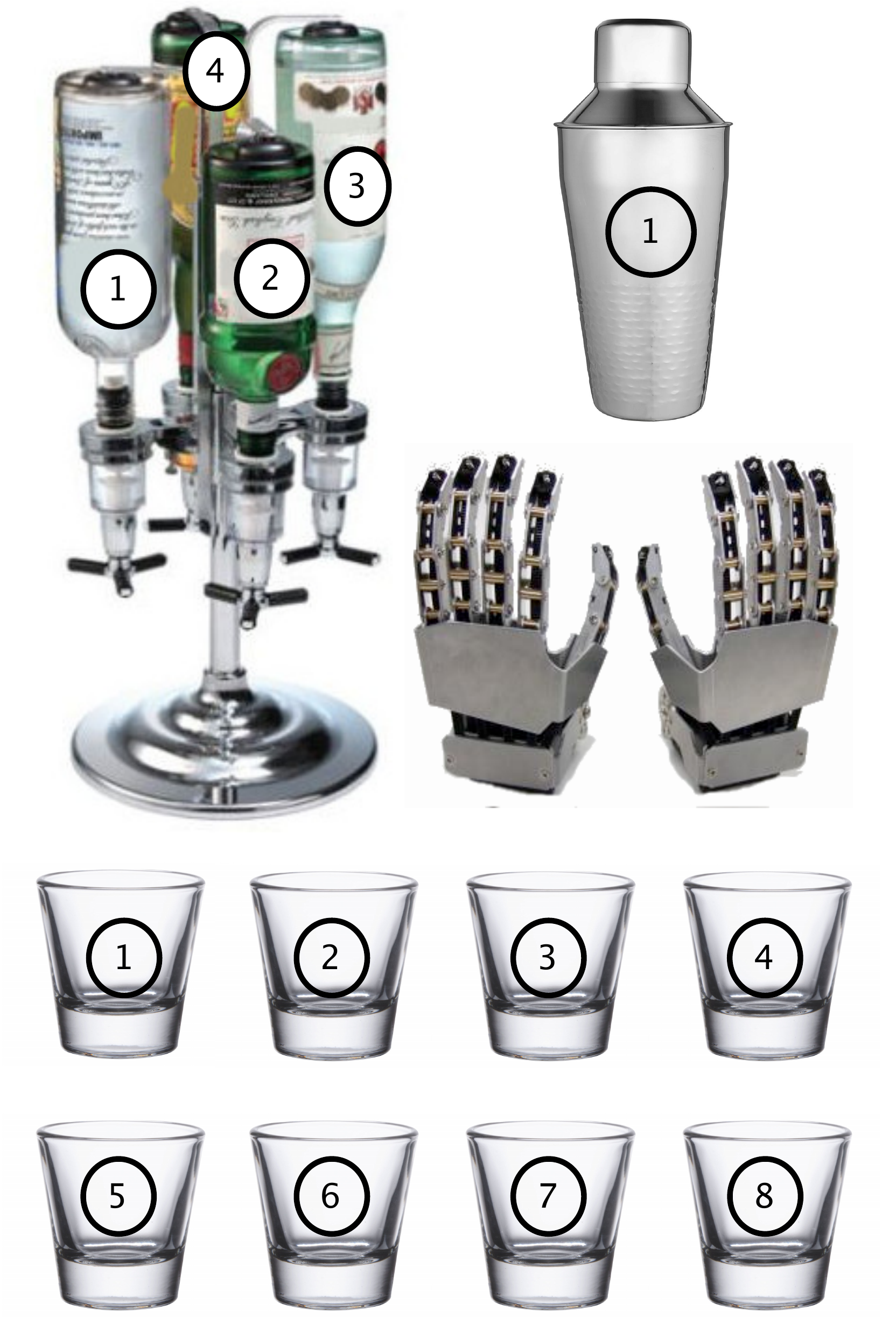}
 \end{tabular}
\caption{\label{fig:barman}Example of  Barman instance}
\end{center}
\end{figure}
In Figure~\ref{fig:barman} we represent the initial configuration and the corresponding input specifications.

Actions available are:
\begin{itemize}
\item {\tt grasp(OBJ)} that executes the grasping either of a specific shot or shaker ({\tt OBJ})
\item {\tt leave(OBJ)} that allows us to leave the shot or shaker  ({\tt OBJ})
\item {\tt fill\_shot(SHOT,ING)} that allows us to fill the shot {\tt SHOT} with the ingredient {\tt ING}
\item {\tt empty\_shot(SHOT)} (resp., {\tt empty\_shaker(SHAKER)}) that allows us  
to empty the shot {\tt SHOT}  (resp., the skaker {\tt SHAKER})

\item {\tt clean\_shot(SHOT)} (resp.,  {\tt clean\_shaker(SHAKER)}) 
that allows us to clean the shot {\tt SHOT}    (resp., the skaker {\tt SHAKER})

\item {\tt pour\_shot\_to\_shaker(SHOT,SHAKER)}
(resp., {\tt  pour\_shaker\_to\_shot(SHAKER,SHOT)}) 
that allows us to pour the content of the shot SHOT in the shaker {\tt SHAKER}
(resp., vice versa).

\item  {\tt shake(SHAKER)} that executes that shaking of the shaker to mix the ingredients.

\item {\tt reduce} (remove a shot from the state once it contains a required cocktail
\end{itemize}
All actions have cost 1 but {\tt reduce} that has cost 0.

\subsection{Cave Diving}

There is a set of divers, each of who can carry four tanks of air. 
These divers must be hired to go into an underwater cave and either take photos or prepare the way for other divers by dropping full tanks of air. The cave is too narrow for more than one diver to enter at a time. 
Divers have a single point of entry. Certain rooms of the cave branches are objectives that the divers must photograph. Swimming and photographing both consume air tanks. Divers must exit the cave and decompress at the end. They can therefore only make a single trip into the cave. 
Certain divers have no confidence in other divers and will refuse to work if someone they have no confidence in has already worked. Divers have hiring costs inversely proportional to how hard they are to work with. This domain was proposed by Nathan Robinson, Christian Muise, and Charles Gretton.

\begin{figure}[htb]
\begin{center}
\begin{tabular}{cc}
{\tt \small
\begin{tabular}[b]{lll}
\%\% Divers information \\
 available(d0)  & available(d1)   & available(d2) \\
 capacity(d0,four)  & capacity(d1,four)  & capacity(d2,four) \\
=(hiring\_cost(d0),60)   & =(hiring\_cost(d1),10) & =(hiring\_cost(d2),10) \\
 precludes(d1,d2) \\
\%\% Cave and tank information\\
 in\_storage(t1)  \\
 next\_tank(t1,t2)   & \dots   & next\_tank(t8,t9) \\
 cave\_entrance(l0)  \\
 connected(l0,l1),  &  \dots &  connected(l5,l1)  \\
\%\%GOAL\\
have\_photo(l4)         & have\_photo(l5) \\
decompressing(d0)   & decompressing(d1) \\
decompressing(d2)   & decompressing(d3) 
\end{tabular}} \\
\includegraphics[width=6cm]{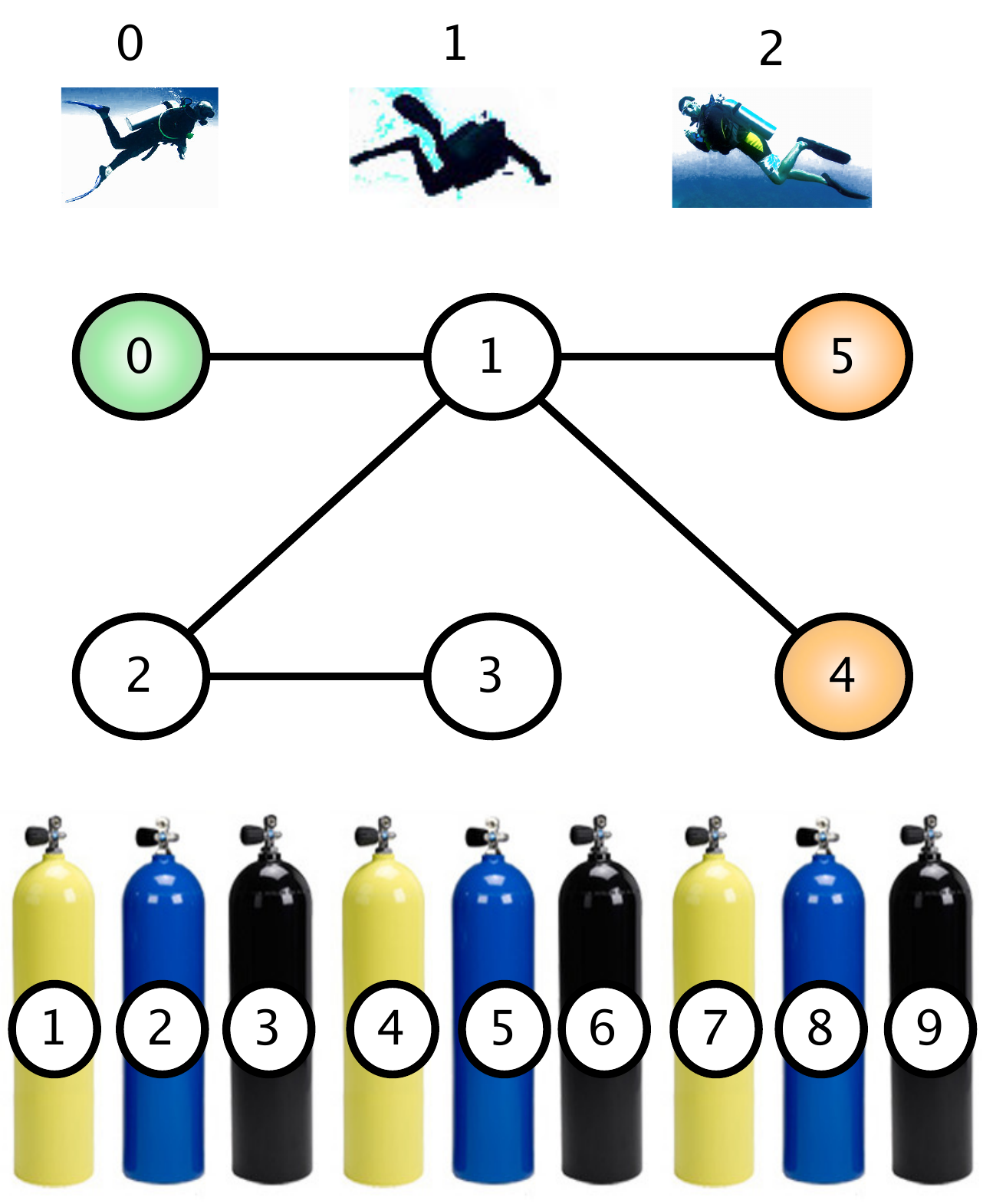} 
\end{tabular}
\caption{\label{fig:cave}Example of Cave Diving instance}
\end{center}
\end{figure}

The cave system is represented by an undirected acyclic graph. 
Divers can carry an amount of tanks according to their capacity.
Rooms that need to be reached are among the leaves of the graph.
In Figure~\ref{fig:cave} we represent an instance of the problem.

The actions available are:
\begin{itemize}
\item {\tt  hire\_diver(Diver)} that requires the availability of hiring cost and should satisfy the
compatibility constraints among divers,

\item {\tt  prepare\_tank(T)} that prepares the tank {\tt T} for the current diver if his capacity allows it,

\item {\tt enter\_water} that requires the diver to be in the cave entrance,

\item {\tt photograph(Loc)} that requires the diver to be in the target location {\tt Loc},

\item {\tt drop\_tank(Loc)} that allows the diver to leave a tank in the location {\tt Loc} (the tank can be either full or empty),

\item {\tt swim(Loc1,Loc2)} that allows the diver to swim between two locations that are adjacent in the graph,

\item {\tt pickup\_tank(Loc)}  that allows the diver to collect a tank stored in the location {\tt Loc},

\item  {\tt decompress} should be made at the end of diving in the cave entrance.

\end{itemize} 
Each action {\tt swim} and {\tt photograph} consumes (empties) one air tank.
All actions but the first one have unitary cost.

\subsection{Childsnack}

This domain is to plan how to make and serve sandwiches for a group of children in which some are allergic to gluten. There are two actions for making sandwiches from their ingredients. The first one makes a sandwich and the second one makes a sandwich taking into account that all ingredients are gluten-free. There are also actions to put a sandwich on a tray and to serve sandwiches. 
Problems in this domain define the ingredients to make sandwiches at the initial state. 
Goals consist of having selected kids served with a sandwich to which they are not allergic. This domain was proposed by Raquel Fuentetaja, Tom\`as de la Rosa Turbides.

\begin{figure}[htb]
\begin{center}
\begin{tabular}{cc}
{\tt\small
\begin{tabular}[b]{lll}
\%\% Positions\\
at(tray1,kitchen)     &      
at(tray2,kitchen) \\
at\_kitchen\_bread(bread1)  &
at\_kitchen\_bread(bread2) \\
at\_kitchen\_bread(bread3)  \\
at\_kitchen\_content(content1)  &
at\_kitchen\_content(content2)  \\
at\_kitchen\_content(content3)  &
at\_kitchen\_content(content4) \\
at\_kitchen\_content(content5)  \\
no\_gluten\_bread(bread3) \\
no\_gluten\_content(content1)  &
no\_gluten\_content(content4) \\
        waiting(child1,table1)  &
        waiting(child2,table2)  \\
        waiting(child3,table3)  &
        waiting(child4,table4)  \\
\%\% Info on allergies\\
allergic\_gluten(child2) &
allergic\_gluten(child4) \\
\%\% Not yet ready sandwhiches\\ 
       notexist(sandw1)  &         notexist(sandw2)  \\
       notexist(sandw3)   & notexist(sandw4) \\
       notexist(sandw5)   & notexist(sandw6) \\
\%\% Goal\\
        served(child2)        &  served(child3)  
\end{tabular}
}\\
\includegraphics[width=6cm]{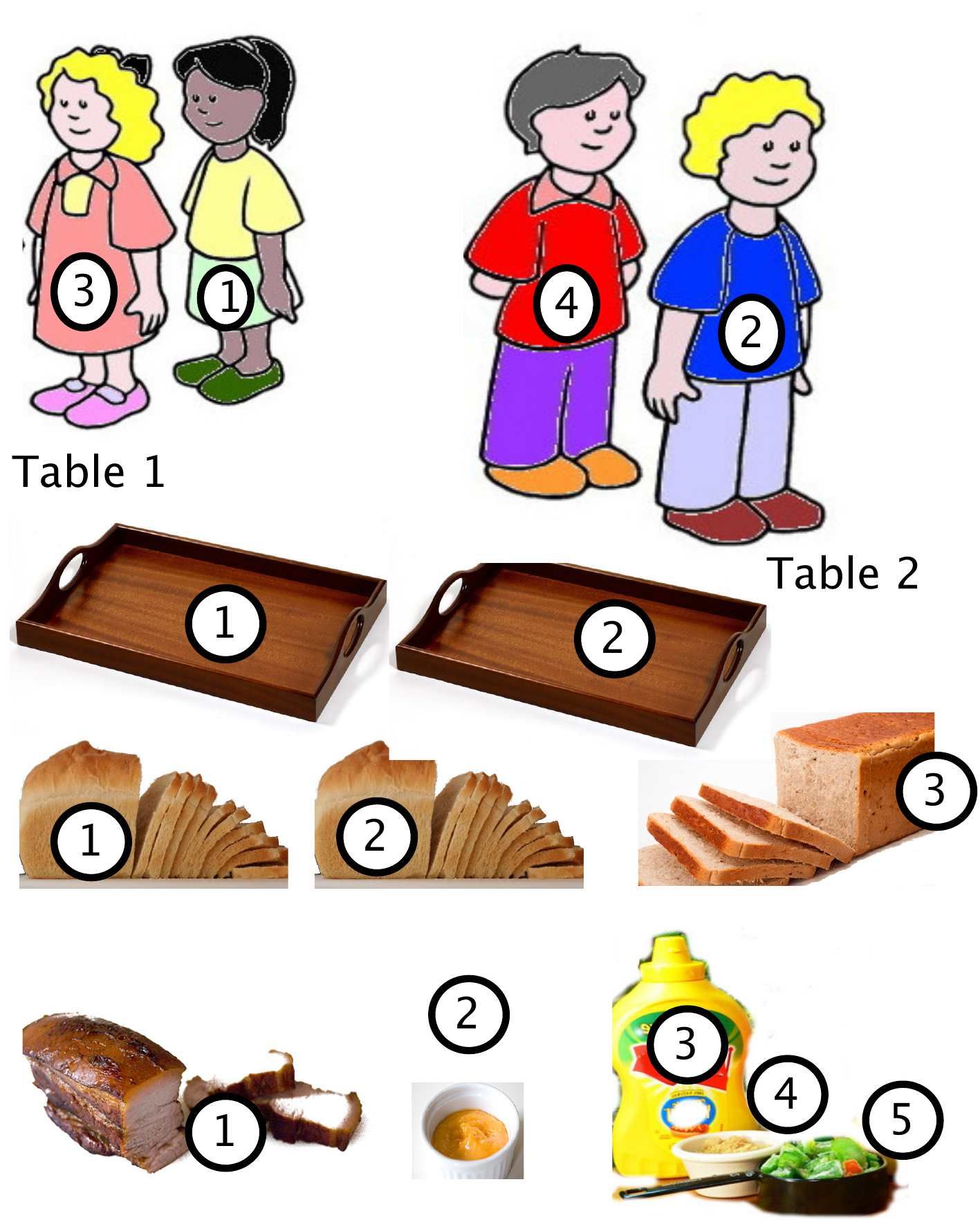} 
\end{tabular}
\caption{\label{fig:snack}Example of Childsnack instance}
\end{center}
\end{figure}

Available actions are the following:

\begin{itemize}
\item {\tt  make\_sandwich\_no\_gluten(Sw,B,Co)} and 
{\tt   make\_sandwich(Sw,B,Co)}
where {\tt SW} is a sandwich,  {\tt B} is a (no-gluten) bread, and 
 {\tt Co} is a (no-gluten) content allows us to make the sandwiches.

 \item {\tt put\_on\_tray(Sw,T)} puts the sandwich {\tt Sw} on the tray {\tt T}
 
 \item
  {\tt serve\_sandwich\_no\_gluten(Sw,Ch,T,Loc)} and
  {\tt serve\_sandwich(Sw,Ch,T,Loc)}
  serves the (no-gluten) sandwich {\tt Sw} which is on tray {\tt T} to the children {\tt Ch} at the location {\tt Loc}

 \item {\tt move\_tray(T,Loc1,Loc2)}, where {\tt T} is a tray, {\tt Loc1} and {\tt Loc2} is a location
 (i.e., a table, the kitchen)
 \end{itemize}
 
 Each action has cost 1. {\tt make\_sandwich} (no-gluten) consumes ingredients.

\subsection{Citycar}
This model aims to simulate the impact of road building/demolition on traffic flows. A city is represented as an acyclic graph, in which each node is a junction and edges are ``potential'' roads. Some cars start from different positions and have to reach their final destination as soon as possible. The agent has a finite number of roads available, which can be built for connecting two junctions and allowing a car to move between them. Roads can also be removed, and placed somewhere else, if needed. In order to place roads or to move cars, the destination junction must be clear, i.e., no cars should be in there. The domain was proposed by Mauro Vallati.

\begin{figure}[htb]
\begin{center}
\begin{tabular}{cc}
{\tt\small
\begin{tabular}[b]{ll}
        \%\% Initial State\\
        same\_line(junction0\_0,junction0\_1)  &         same\_line(junction0\_1,junction0\_0)\\
        same\_line(junction1\_0,junction1\_1) &         same\_line(junction1\_1,junction1\_0)\\
        same\_line(junction0\_0,junction1\_0) &        same\_line(junction1\_0,junction0\_0)\\
        same\_line(junction0\_1,junction1\_1) &        same\_line(junction1\_1,junction0\_1)\\
        diagonal(junction0\_0,junction1\_1) &        diagonal(junction1\_1,junction0\_0)\\
        diagonal(junction0\_1,junction1\_0) &        diagonal(junction1\_0,junction0\_1)\\
        clear(junction0\_0) &
        clear(junction0\_1)\\
        clear(junction1\_0) &
        clear(junction1\_1)\\
        at\_garage(garage0,junction0\_1)\\
        starting(car0,garage0) &
        starting(car1,garage0)\\
        \%\% GOAL\\
        arrived(car0,junction1\_1) &
        arrived(car1,junction1\_0)
\end{tabular}
}\\
\includegraphics[width=8cm]{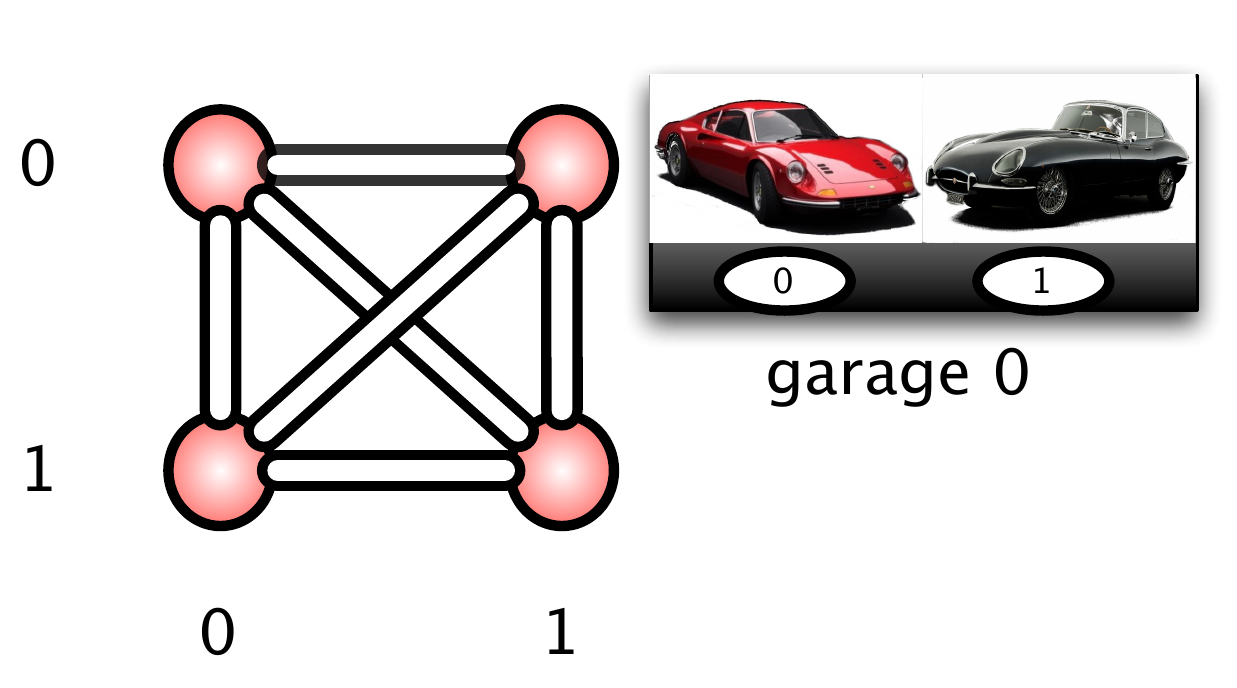} 
\end{tabular}
\caption{\label{fig:city cars}Example of Citycars instance}
\end{center}
\end{figure}

Allowed actions are the following:

\begin{itemize}
\item {\tt car\_arrived(Dest)}, which has cost 0. It allows to remove a car
from the network and to remove the occurrence of the destination {\tt Dest} 
(a junction)
from the list of all final destinations.

\item {\tt car\_start(Loc)}: A car is put in the road from the garage of location {\tt Loc}: it has cost 1.

\item {\tt move\_car\_in\_road(FromLoc)} allows us to move a car in a road 
from the junction {\tt FromLoc} (cost 1---the road is a straight line or a diagonal road starting
in {\tt FromLoc}).

\item {\tt   move\_car\_out\_road(ToLoc)} allows us to move a  out of a road 
as soon as the junction {\tt ToLoc} is reached by the car (cost 1---the 
road is a straight line or a diagonal road ending
in {\tt ToLoc}).

\item These actions allow us to build diagonal, straight roads or of deleting one road:
\begin{itemize}
\item
{\tt build\_diagonal\_oneway(FromLoc,ToLoc)} (cost 30), 
\item
{\tt  build\_straight\_oneway(FromLoc,ToLoc)} (cost 20),
\item
{\tt  destroy\_road(FromLoc,ToLoc)} (cost 10).
\end{itemize}

\end{itemize}

Let us observe that search symmetries are eliminated by considering the cars equivalent during the search.
It is trivial to label them a-posteriori given a correct plan.

\subsection{GED}
The GED problem is to find a min-cost sequence of operations that transforms one genome (signed permutation of genes) into another. The purpose of this is to use this cost as a measure of the distance between the two genomes, which is used to construct hypotheses about the evolutionary relationship between the organisms. The domains was proposed by Patrik Haslum.

\medskip

This problem can be stated at several abstraction levels. 
A general version could include gene insertions and deletions. 
Let us focus on the abstraction level and on the three rules
required by the competition benchmarks.   

A gene is identified by a symbolic name.
The connection between genes is stated by a binary predicate {\tt cw} that 
encodes a linear graph. 
Each gene can occur in a regular direction (normal) or in reverse direction
(inverted).

The three rules allowed are cut (of a substring) from the main genome, and then a splice of the cut substring directed or reversed in a selected point of the main genome. The reverse of a single gene is also allowed.
Just to fix the ideas, let us consider the example in figure \ref{fig:GED}. 
Reversed genes are overlined.

\begin{figure}[h]
\hspace*{-.6\textwidth}\begin{tabular}{p{.3\textwidth}p{.4\textwidth}p{.22\textwidth}}
{\tt\small
\begin{tabular}{p{.3\textwidth}}
\%\% INITIAL STATE\\
\\
        normal(a),\\
        normal(b),\\
        normal(c),\\
        normal(d),\\
        cw(a,b),\\
        cw(b,c),\\
        cw(c,d)
 \end{tabular}
 }
 &
 \begin{tabular}{p{.3\textwidth}}
 $\mathtt{a}\cdot \mathtt{b} \cdot \mathtt{c} \cdot \mathtt{d}$\\
 $\Downarrow$ (cut 2--4, temp situation)\\
 $\mathtt{a} \phantom{aaaaaa} \mathtt{b}  \cdot \mathtt{c} \phantom{aaaaaa} \mathtt{d}$\\
 $\Downarrow$ (cut 2-4, final situation)\\
 $\mathtt{a} \cdot \mathtt{d} \phantom{aaaaaa} \mathtt{b}  \cdot \mathtt{c} $\\
  $\Downarrow$ (reverse of the 2nd string)\\
 $\mathtt{a}\cdot  \mathtt{d} \phantom{aaaaaa}\overline{\mathtt{c}} \cdot \overline{\mathtt{b}}$\\
  $\Downarrow$ (and  splice in the 1st)\\
 $\mathtt{a}\cdot \overline{\mathtt{c}} \cdot \overline{\mathtt{b}} \cdot \mathtt{d}$\\
\end{tabular}
&
{\tt\small
\begin{tabular}{p{.22\textwidth}}
\%\% GOAL\\
\\
        normal(a),\\
        inverted(b),\\
        inverted(c),\\
        normal(d),\\
        cw(a,c),\\
        cw(c,b),\\
        cw(b,d)\\
\end{tabular}
}
\end{tabular}
\caption{\label{fig:GED}An instance of the GED problem and a possible solution}
\end{figure}

Each complex action (cut and splice) is split in some sub-actions as done by 
Patrik Haslum in his PDDL encoding (\url{http://picat-lang.org/ipc14/ged.pddl}).

\subsection{Floortile, Parking, and Tetris}
\begin{figure}[ht]
\begin{center}
\begin{tabular}{c}
{\tt \small
\begin{tabular}[b]{ll}
\%\% Floor description\\
clear(01) & \dots  \phantom{aa} clear(64),\\
up(11,01) & \dots \phantom{aa}   up(64,54),\\
down(01,11)  & \dots \phantom{aa}    down(54,64) \\
right(02,01) & \dots  \phantom{aa}   right(64,63) \\
left(01,02) & \dots  \phantom{aa}   left(63,64)\\ 
\%\% Robots positions and states\\
robot\_at(robot1,11) &           robot\_has(robot1,white) \\
robot\_at(robot2,52) &           robot\_has(robot2,black)\\
available\_color(white) &        available\_color(black)\\
\%\% GOAL\\
painted(11,white) &  painted(12,black) \\   painted(13,white) &
        painted(14,black) \\
$\vdots$ & $\vdots$\\
 painted(61,black)  &
        painted(62,white) \\
        painted(63,black) &
        painted(64,white) 
 \end{tabular}
 }\\
\includegraphics[width=3cm]{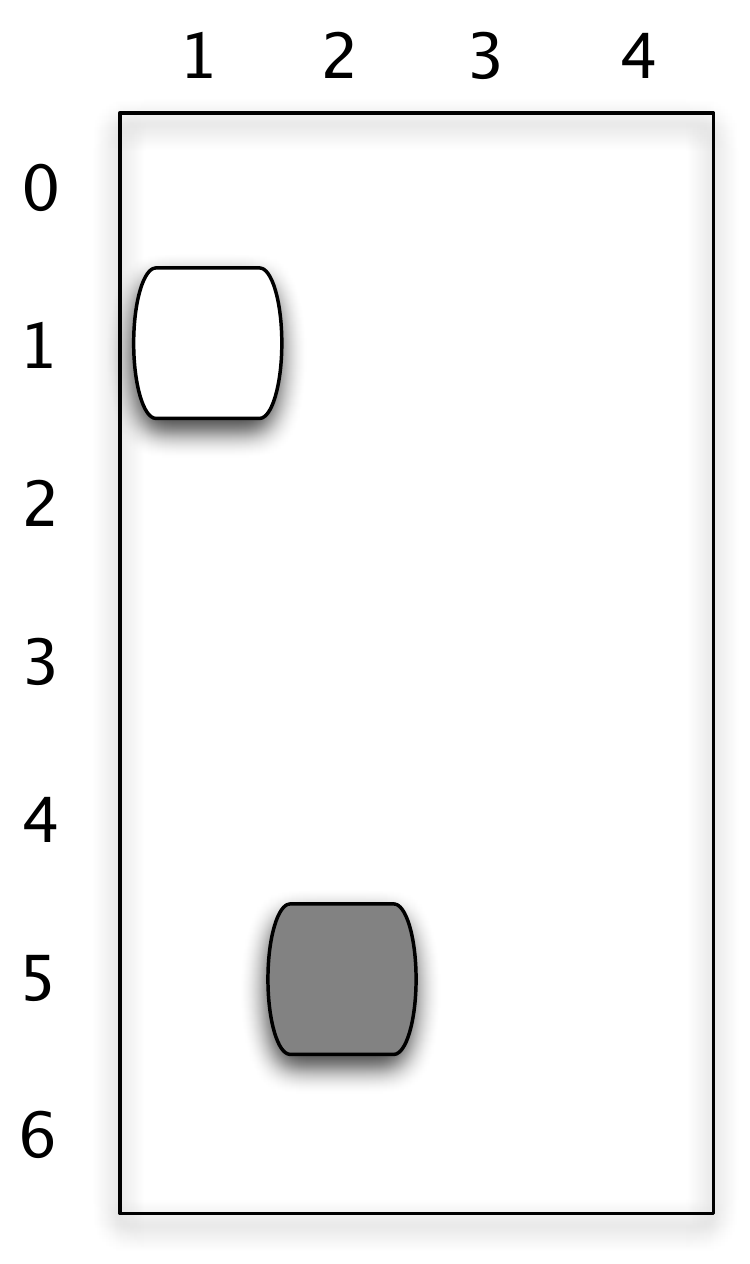} 
\phantom{aaa}
\includegraphics[width=3cm]{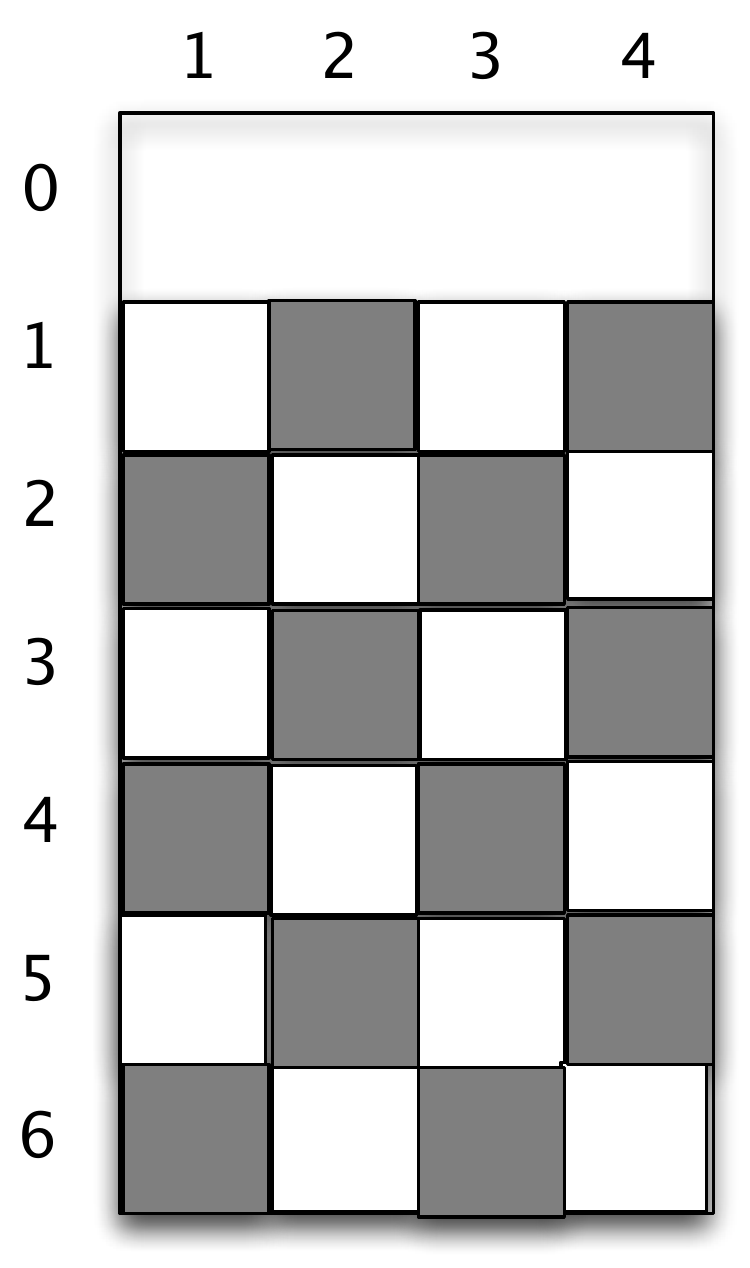} 
\end{tabular}
\caption{\label{fig:floortile}Example of Floortile instance. A solution with plancost 104 exists (benchmark instance {\tt p01642}). }
\end{center}
\end{figure}

\begin{figure}[ht]
\begin{center}
\begin{tabular}{c}
{\tt \small
\begin{tabular}[b]{ll}
\%\% INITIAL STATE\\
\\
        at\_curb(car3), & 
        at\_curb\_num(car3,curb0), \\ 
        behind\_car(car2,car3), &
        car\_clear(car2), \\
        at\_curb(car4), & 
        at\_curb\_num(car4,curb1), \\ 
        behind\_car(car10,car4), & 
        car\_clear(car10), \\
        at\_curb(car0), & 
        at\_curb\_num(car0,curb2), \\ 
        behind\_car(car5,car0), & 
        car\_clear(car5), \\
        at\_curb(car1), &
        at\_curb\_num(car1,curb3), \\
        behind\_car(car9,car1), &
        car\_clear(car9),\\
        at\_curb(car7), & 
        at\_curb\_num(car7,curb4), \\
        behind\_car(car8,car7), &
        car\_clear(car8), \\
        at\_curb(car11), &
        at\_curb\_num(car11,curb5), \\
        behind\_car(car6,car11), &
        car\_clear(car6),\\
        curb\_clear(curb6)\\ 
\\        
\%\% GOAL\\
\\
        at\_curb\_num(car0,curb0), &
        behind\_car(car7,car0), \\
        at\_curb\_num(car1,curb1), &
        behind\_car(car8,car1), \\
        at\_curb\_num(car2,curb2), &
        behind\_car(car9,car2), \\
        at\_curb\_num(car3,curb3), &
        behind\_car(car10,car3), \\
        at\_curb\_num(car4,curb4), &
        behind\_car(car11,car4), \\ 
        at\_curb\_num(car5,curb5), &
        at\_curb\_num(car6,curb6) \\
 \end{tabular}
 }
 \\
 \includegraphics[width=.2\textwidth]{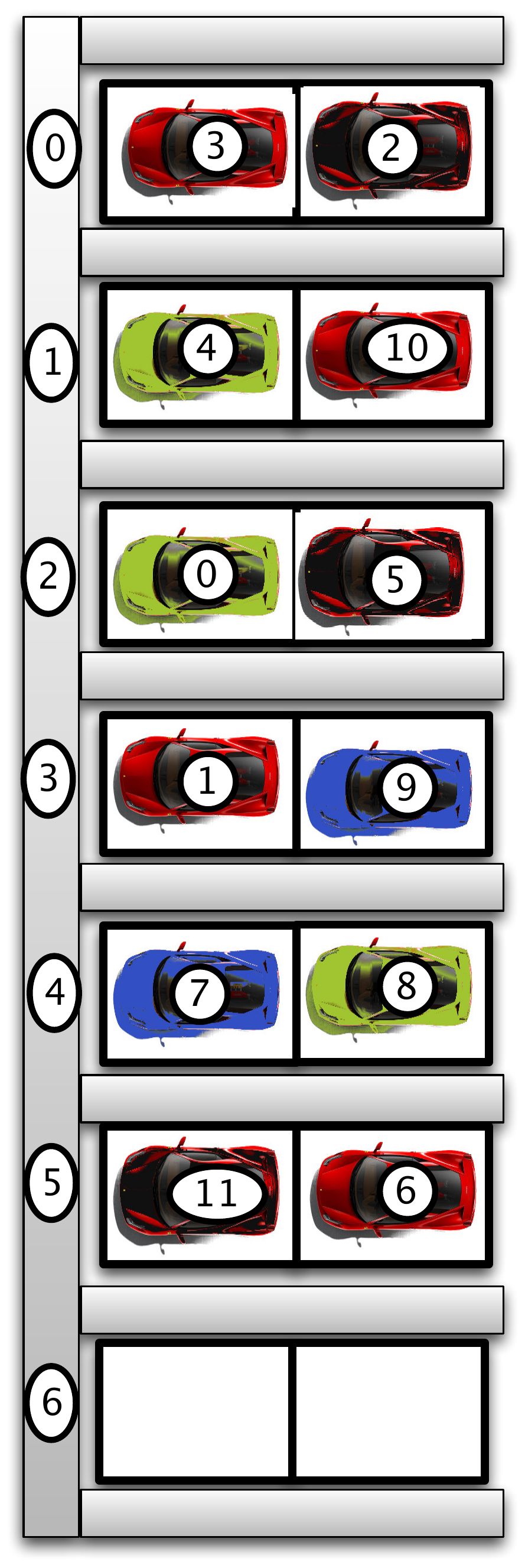} \phantom{aaaaaaaaa}
 \includegraphics[width=.2\textwidth]{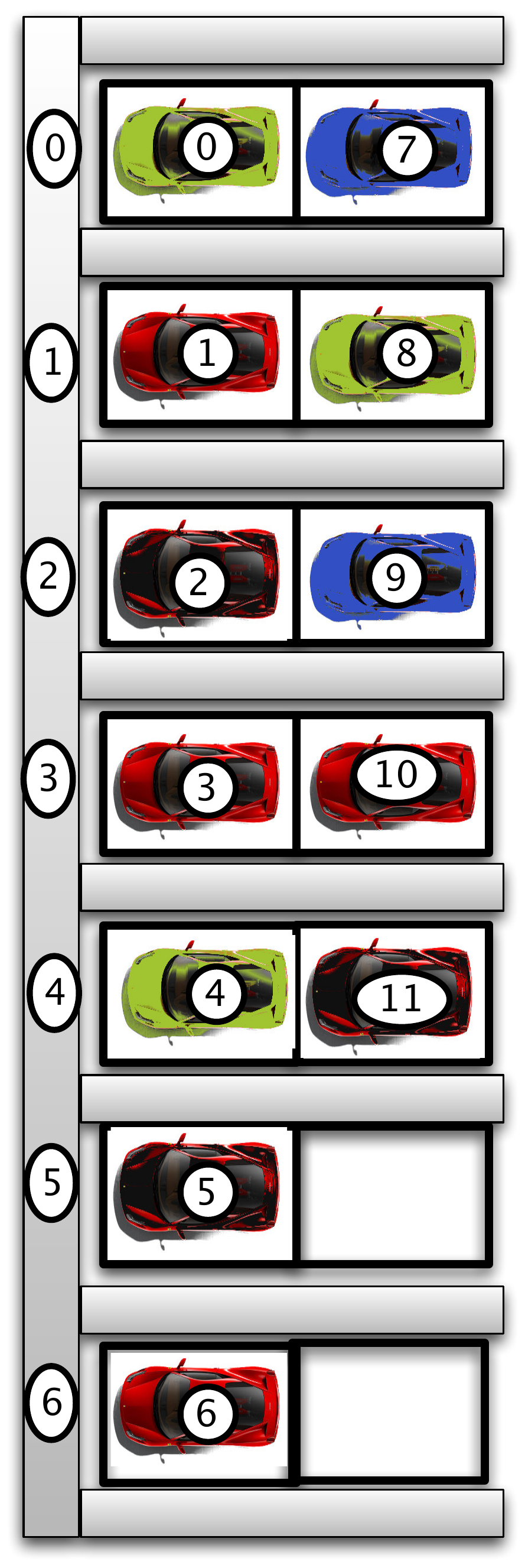}
 \end{tabular}   
 \end{center}
 
\caption{\label{fig:parking}
An instance of parking (left: initial state, right: goal). 
A solution with 18 moves exists 
(benchmark instance {\tt p\_12\_7\_01}).}
\end{figure}

\begin{figure}[ht]
\begin{center}
\begin{tabular}{cc}
{\tt \small 
\begin{tabular}[b]{llll}
\%\% Board description\\
     connected(f0\_0f,f0\_1f), & \dots &   connected(f0\_2f,f0\_3f) \\
    connected(f1\_1f,f1\_0f), & \dots &   connected(f1\_2f,f1\_3f), \\
   & \dots \\
connected(f6\_0f,f7\_0f), & \dots &  connected(f6\_3f,f7\_3f)\\
clear(f0\_3f), & \dots &   clear(f7\_3f), \\
\%\% Pieces\\
        at\_right\_l(rightl0,f0\_0f,f1\_0f,f1\_1f), &&
        at\_right\_l(rightl1,f2\_1f,f3\_1f,f3\_2f),\\
        at\_two(straight0,f0\_2f,f1\_2f),&&
        at\_square(square0,f0\_1f)\\
\%\% Goal \\
clear(f0\_0f), & \dots & clear(f0\_3f)\\
clear(f1\_0f)  & \dots & clear(f1\_3f)\\
clear(f2\_0f)  & \dots & clear(f2\_3f)\\
clear(f3\_0f) & \dots & clear(f3\_3f)
 \end{tabular}
 }\\
\includegraphics[width=.2\textwidth]{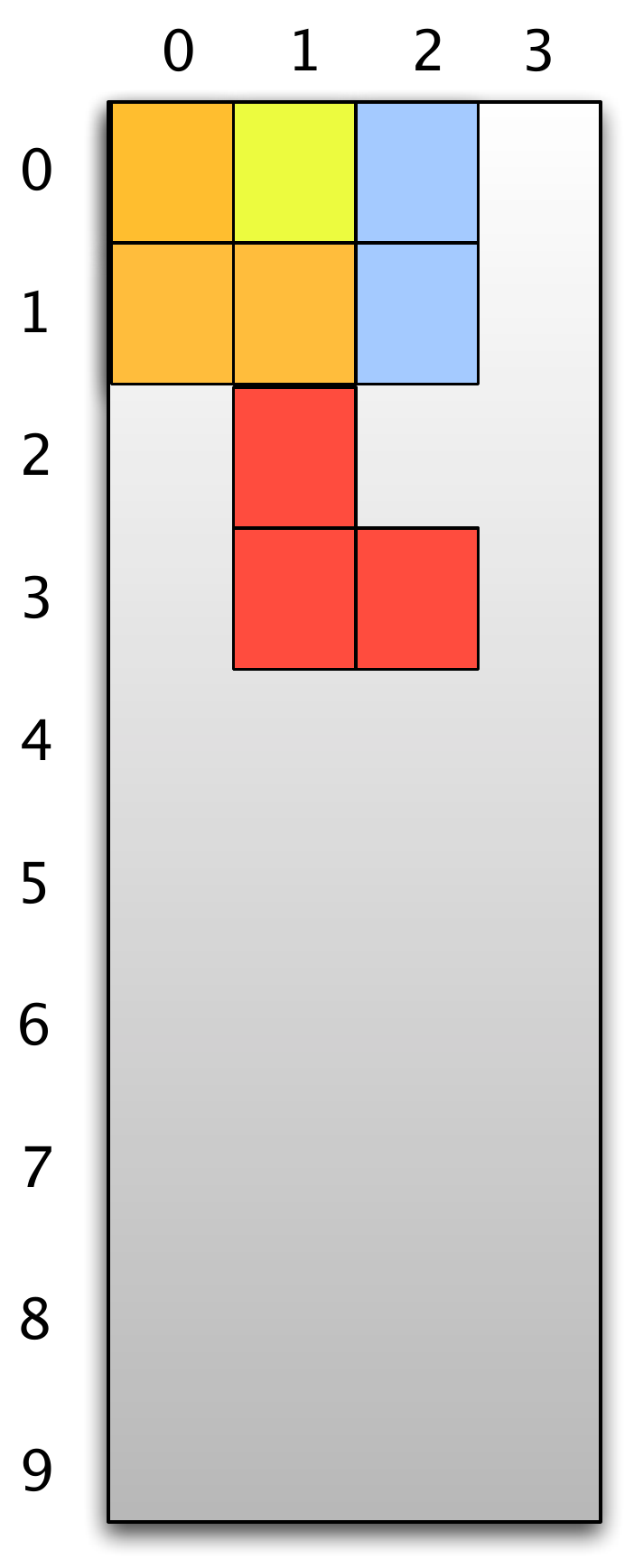} 
\includegraphics[width=.2\textwidth]{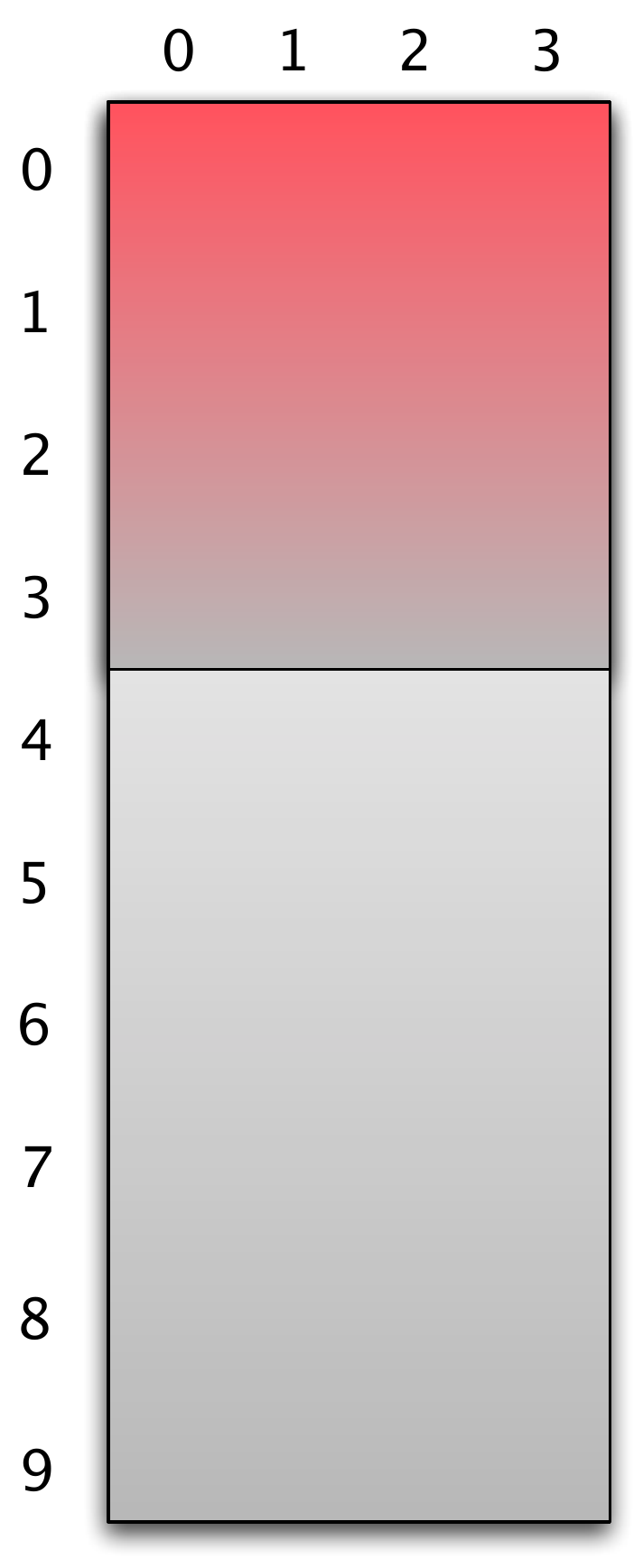} 
\includegraphics[width=.2\textwidth]{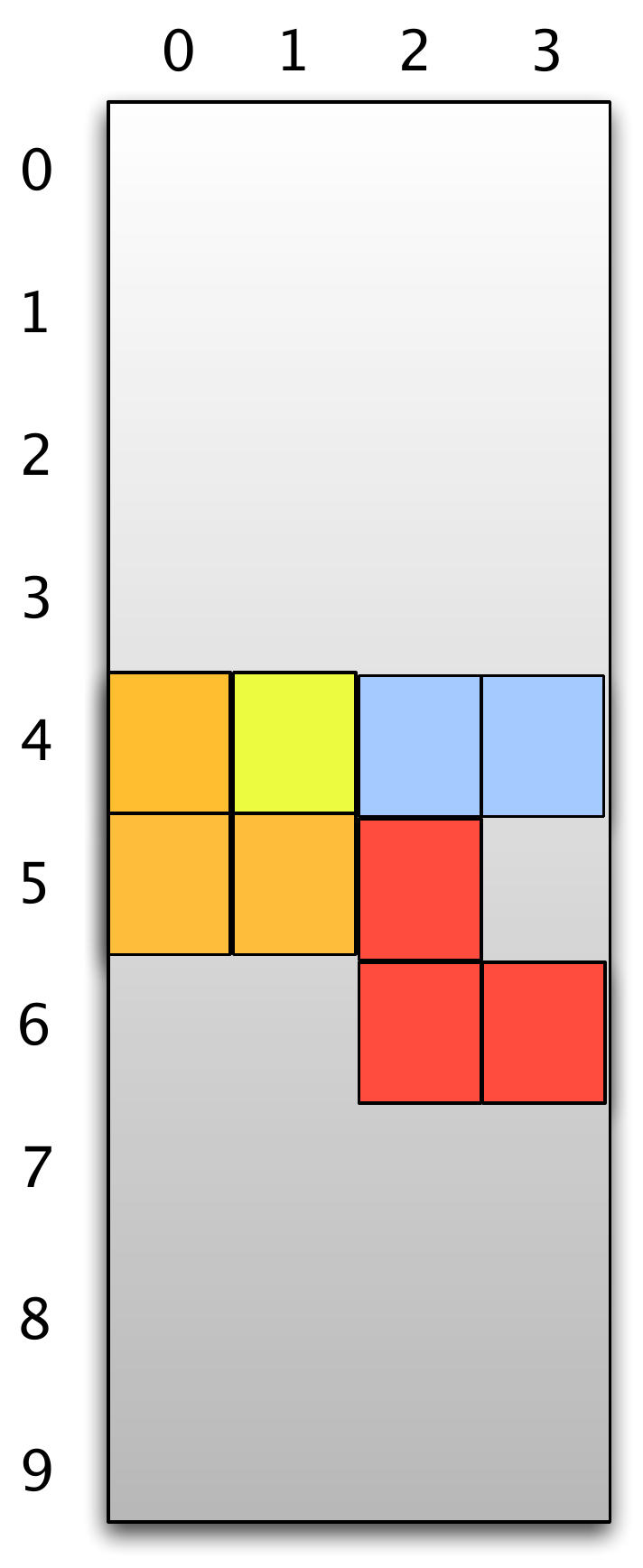} 
 \end{tabular}
\caption{\label{fig:tetris}Example of Tetris instance: initial state (left), goal (center). 
A plan of length 36 exists (instance {\tt 01\_8} of the benchmarks) 
leading to the final situation to the right.}
\end{center}
\end{figure}

For the three domains discussed extensively in the core of paper we only show here an instance both in concrete form and as a picture (see Figures \ref{fig:floortile}--\ref{fig:tetris}). The Transport domain is discussed in detail in the next section.

\clearpage
\section{The Transport Domain}
\subsection{PDDL Encoding of the Transport Domain}
\begin{scriptsize}
\begin{verbatim}
(define (domain transport)
  (:requirements :typing :action-costs)
  (:types
        location target locatable - object
        vehicle package - locatable
        capacity-number - object
  )

  (:predicates 
     (road ?l1 ?l2 - location)
     (at ?x - locatable ?v - location)
     (in ?x - package ?v - vehicle)
     (capacity ?v - vehicle ?s1 - capacity-number)
     (capacity-predecessor ?s1 ?s2 - capacity-number)
  )

  (:functions
     (road-length ?l1 ?l2 - location) - number
     (total-cost) - number
  )

  (:action drive
    :parameters (?v - vehicle ?l1 ?l2 - location)
    :precondition (and
        (at ?v ?l1)
        (road ?l1 ?l2)
      )
    :effect (and
        (not (at ?v ?l1))
        (at ?v ?l2)
        (increase (total-cost) (road-length ?l1 ?l2))
      )
  )

 (:action pick-up
    :parameters (?v - vehicle ?l - location ?p - package ?s1 ?s2 - capacity-number)
    :precondition (and
        (at ?v ?l)
        (at ?p ?l)
        (capacity-predecessor ?s1 ?s2)
        (capacity ?v ?s2)
      )
    :effect (and
        (not (at ?p ?l))
        (in ?p ?v)
        (capacity ?v ?s1)
        (not (capacity ?v ?s2))
        (increase (total-cost) 1)
      )
  )

  (:action drop
    :parameters (?v - vehicle ?l - location ?p - package ?s1 ?s2 - capacity-number)
    :precondition (and
        (at ?v ?l)
        (in ?p ?v)
        (capacity-predecessor ?s1 ?s2)
        (capacity ?v ?s1)
      )
    :effect (and
        (not (in ?p ?v))
        (at ?p ?l)
        (capacity ?v ?s2)
        (not (capacity ?v ?s1))
        (increase (total-cost) 1)
      )
  )

)
\end{verbatim}
\end{scriptsize}

\subsection{Picat Encoding of the Transport Domain}
\begin{scriptsize}
\begin{verbatim}
final({Trucks,[]}) =>   % no waiting packages and no loaded packages
    foreach([_Loc,Dests|_] in Trucks)
        Dests == []
    end.

% unload a package
action({Trucks,Packages},NextState,Action,ActionCost),
    select([Loc,Dests,Cap],Trucks,TrucksR),
    select(Loc,Dests,DestsR)   % unload it deterministically 
=>
    Action = $unload(Loc),    
    ActionCost = 1,
    NewTrucks = insert_ordered(TrucksR,[Loc,DestsR,Cap]),
    NextState = {NewTrucks,Packages}.
action({Trucks,Packages},NextState,Action,ActionCost) ?=>
    Action = $unload(Loc),    
    ActionCost = 1,
    select([Loc,Dests,Cap],Trucks,TrucksR),
    select(Dest,Dests,DestsR),
    NewTrucks = insert_ordered(TrucksR,[Loc,DestsR,Cap]),
    NewPackages = insert_ordered(Packages,(Loc,Dest)),
    NextState = {NewTrucks,NewPackages}.

% load a package onto a truck if the truck and the package are at the same location
action({Trucks,Packages},NextState,Action,ActionCost) ?=>
    Action = $load(Loc),
    ActionCost = 1,
    select([Loc,Dests,Cap],Trucks,TrucksR),
    length(Dests) < Cap,
    select((Loc,Dest),Packages,PackagesR),   % the package is at the same location as the truck
    NewTrucks = insert_ordered(TrucksR,[Loc,insert_ordered(Dests,Dest),Cap]),
    NextState = {NewTrucks,PackagesR}.

% drive a truck from Loc to NextLoc
action({Trucks,Packages},NextState,Action,ActionCost) =>
    Action = $move(Loc,NextLoc),
    select([Loc|Tail],Trucks,TrucksR),
    road(Loc,NextLoc,ActionCost),
    NewTrucks = insert_ordered(TrucksR,[NextLoc|Tail]),
    NextState = {NewTrucks,Packages},
    estimate_cost(NextState) =< current_resource()-ActionCost.

table
estimate_cost({Trucks,Packages}) = Cost =>
    LoadedPackages = [(Loc,Dest) : [Loc,Dests,_] in Trucks, Dest in Dests],
    NumLoadedPackages = length(LoadedPackages),
    TruckLocs = [Loc : [Loc|_] in Trucks],
    travel_cost(TruckLocs,LoadedPackages,Packages,0,TCost),
    Cost = TCost+NumLoadedPackages+length(Packages)*2.  % includes load and unload costs

% the maximum of the minimum cost of transporting each single package
travel_cost(_Trucks,[],[],Cost0,Cost) => Cost=Cost0.
travel_cost(Trucks,[(PLoc,PDest)|Packages],Packages2,Cost0,Cost) => 
    Cost1 = min([D1+D2 : TLoc in Trucks,
                        shortest_dist(TLoc,PLoc,D1),
                        shortest_dist(PLoc,PDest,D2)]),
    travel_cost(Trucks,Packages,Packages2,max(Cost0,Cost1),Cost).
travel_cost(Trucks,[],Packages2,Cost0,Cost) => 
    travel_cost(Trucks,Packages2,[],Cost0,Cost).
\end{verbatim}
\end{scriptsize}

\clearpage
\subsection{An Instance of the Transport Domain}

\begin{figure}[h]
\begin{center}
\begin{tabular}{cc}
\includegraphics[width=3.0cm]{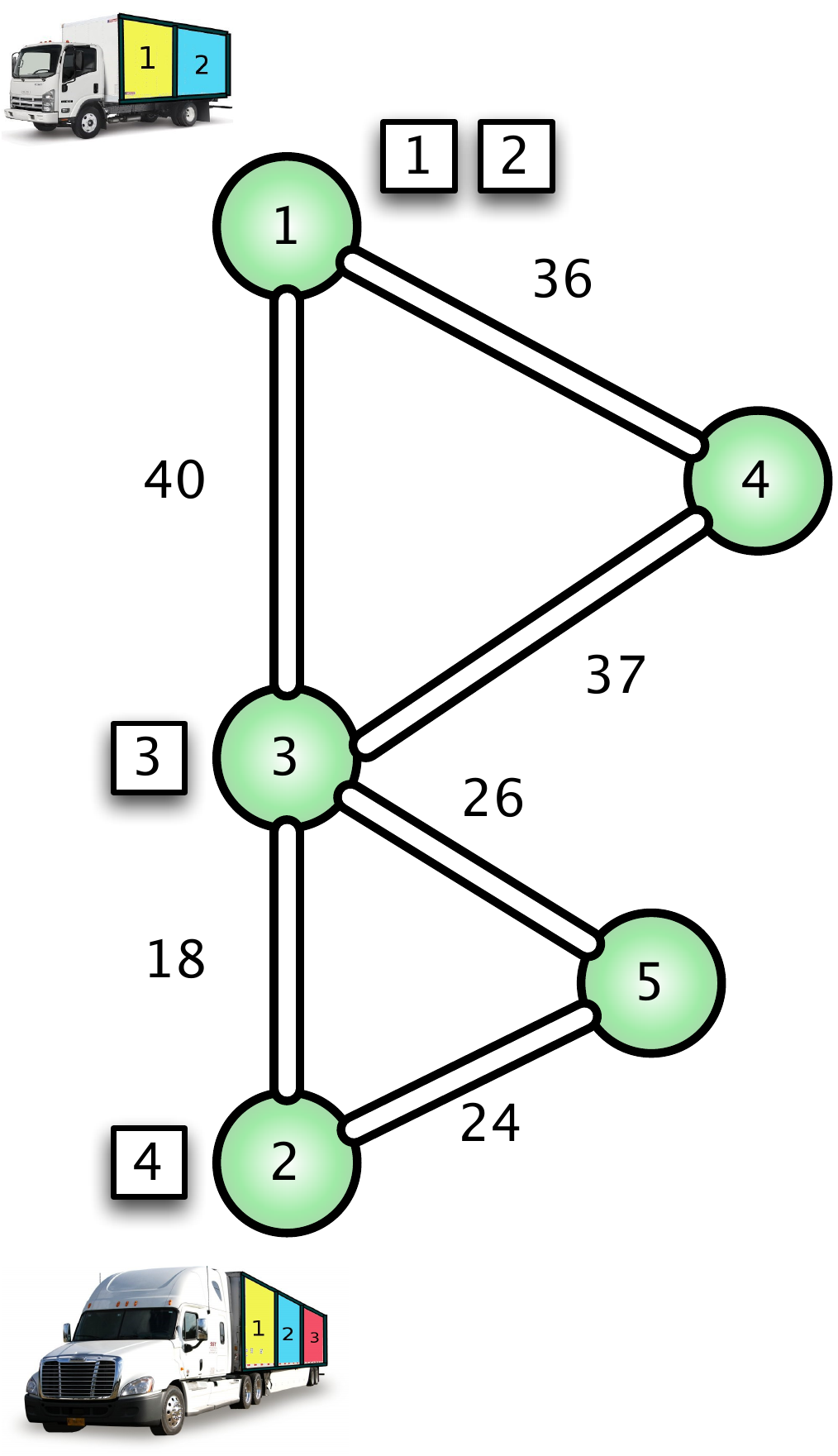} &
\includegraphics[width=3.5cm]{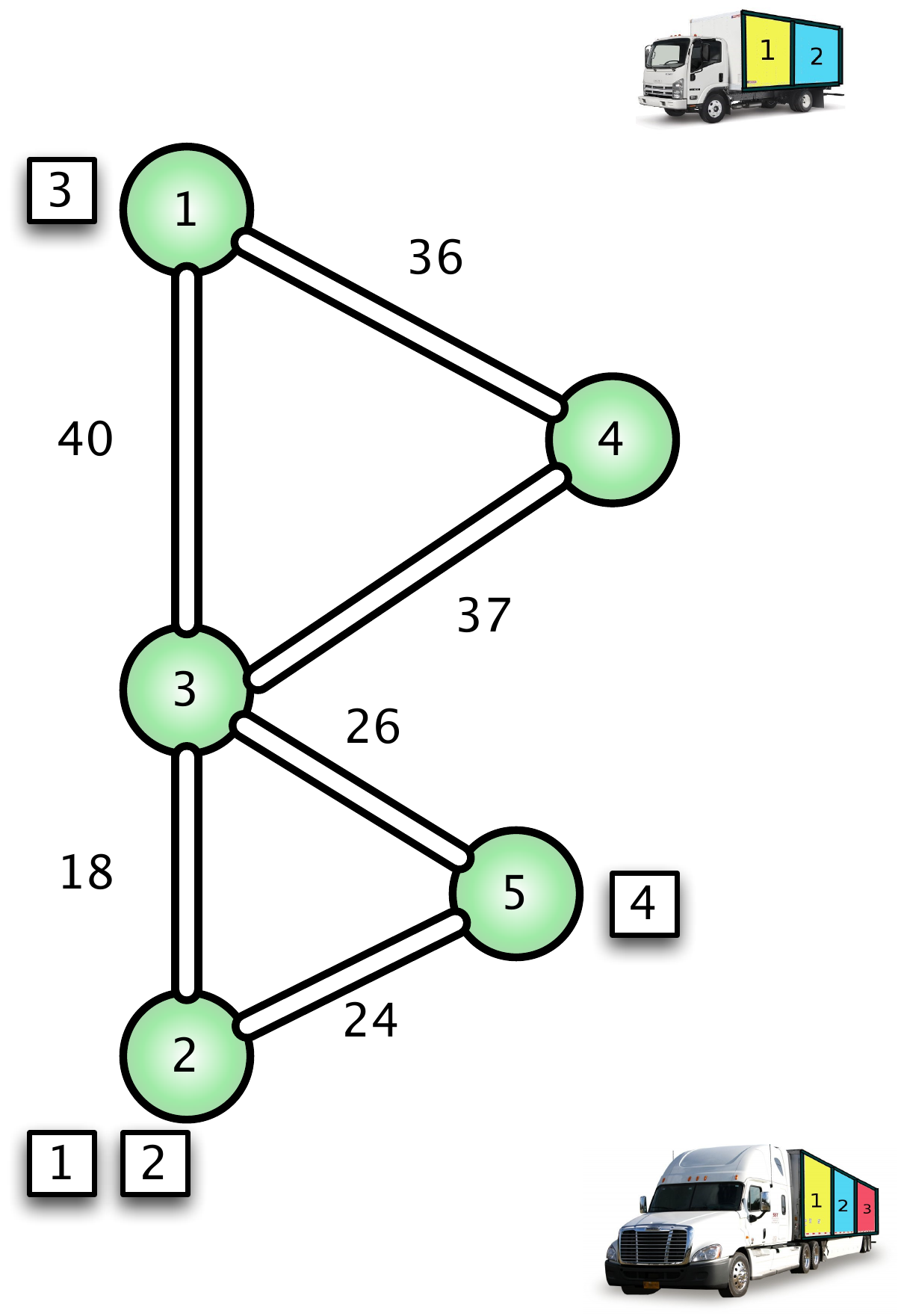}  \\
Initial state & Goal state \\
\end{tabular}
\caption{\label{fig:transport}An Instance of the Transport Domain (p01).}
\end{center}
\end{figure}

\subsubsection{Solving the instance with Picat}
\begin{scriptsize}
\begin{verbatim}
main =>
    Facts = 
      $[road(c3,c1,40),road(c1,c3,40),road(c3,c2,18),
        road(c2,c3,18),road(c4,c1,36),road(c1,c4,36),
        road(c4,c3,37),road(c3,c4,37),road(c5,c2,24),
        road(c2,c5,24),road(c5,c3,26),road(c3,c5,26)],
    cl_facts(Facts,[$road(+,-,-)]),
    Trucks = [[c2,[],3],[c1,[],2]],
    Packages = [(c1,c2),(c1,c2),(c3,c1),(c2,c5)],
    best_plan({sort(Trucks),sort(Packages)},Plan,PlanCost),
    foreach ({I,Action} in zip(1..len(Plan),Plan))
        printf("%3d. %w\n",I,Action)
    end,
    println(plan_cost=PlanCost).
\end{verbatim}
\end{scriptsize}

\subsubsection{An Optimal Plan for the Instance}
\begin{scriptsize}
\begin{verbatim}
  1. load(c1)
  2. load(c1)
  3. load(c2)
  4. move(c1,c3)
  5. move(c2,c5)
  6. unload(c5)
  7. move(c3,c2)
  8. unload(c2)
  9. unload(c2)
 10. move(c2,c3)
 11. load(c3)
 12. move(c3,c1)
 13. unload(c1)

 plan_cost = 148
\end{verbatim}
\end{scriptsize}

\end{document}